 \documentclass[review,3p,times,authoryear]{elsarticle}

\usepackage{amssymb}

\usepackage{amsmath}

\usepackage{float}

\usepackage{adjustbox}

\usepackage{soul}

\usepackage[ruled]{algorithm2e}
\usepackage{setspace}

\usepackage{graphicx}
\usepackage{subfig}

\usepackage{makecell}
\usepackage{multirow}
\usepackage{fontawesome} 

\usepackage{hyperref}
\usepackage{xcolor}
\hypersetup{
    colorlinks=true,
    linkcolor=blue, 
    urlcolor=blue,  
    citecolor=blue, 
}

\usepackage{caption}
\journal{Engineering Applications of Artificial Intelligence}

\begin{document}

\begin{frontmatter}



\title{Train a Real-world Local Path Planner in One Hour via Partially Decoupled Reinforcement Learning and Vectorized Diversity}

\author[1]{Jinghao Xin}
\ead{xjhzsj2019@sjtu.edu.cn}

\author[2,3]{Jinwoo Kim}
\ead{jinwookim@hanyang.ac.kr}

\author[1]{Zhi Li}
\ead{lizhibeaman@sjtu.edu.cn}

\author[1]{Ning Li \corref{cor1}}
\ead{ning\_li@sjtu.edu.cn}

\address[1]{Department of Automation, Shanghai Jiao Tong University, Shanghai 200240, P.R. China}
\address[2]{Department of Civil and Environmental Engineering, Hanyang University, Seoul 04763, Republic of Korea}
\address[3]{School of Civil and Environmental Engineering, Nanyang Technological University, S639798, Singapore}

\cortext[cor1]{Corresponding author}


\begin{abstract}
\textcolor{black}{Deep Reinforcement Learning (DRL) has exhibited efficacy in resolving the Local Path Planning (LPP) problem. However, its practical application remains significantly constrained due to its limited training efficiency and generalization capability.
To address these challenges, we propose a solution termed Color, which includes an Actor-Sharer-Learner (ASL) training framework designed to improve efficiency, and a fast yet diverse simulator named Sparrow aimed at elevating both efficiency and generalization.
Specifically, the ASL employs a Vectorized Data Collection (VDC) mode to enhance data collection, decouples the model optimization from data collection to expedite data consumption, and partially connects the two procedures with a Time Feedback Mechanism (TFM) to evade data underuse or overuse. 
Meanwhile, the Sparrow simulator utilizes a 2-Dimensional (2D) grid-based world, simplified kinematics, matrix operation, and conversion-free data flow to achieve a lightweight design. The lightness facilitates vectorized diversity, allowing for rapid and diversified simulation across numerous copies of the vectorized environments, thereby significantly enhancing both efficiency and generalization capacity.
Comprehensive experiments demonstrate that with merely one hour of simulation training, Color achieves impressive arrival rates of 84\% and 90\% on 32 simulated and 42 real-world LPP scenarios, respectively. 
The code and video of this paper are accessible on our website\footnote{https://github.com/XinJingHao/Color}. }
\end{abstract}




\end{frontmatter}


\begin{figure}[H]
	\centering
	\includegraphics[width=0.76\textwidth]{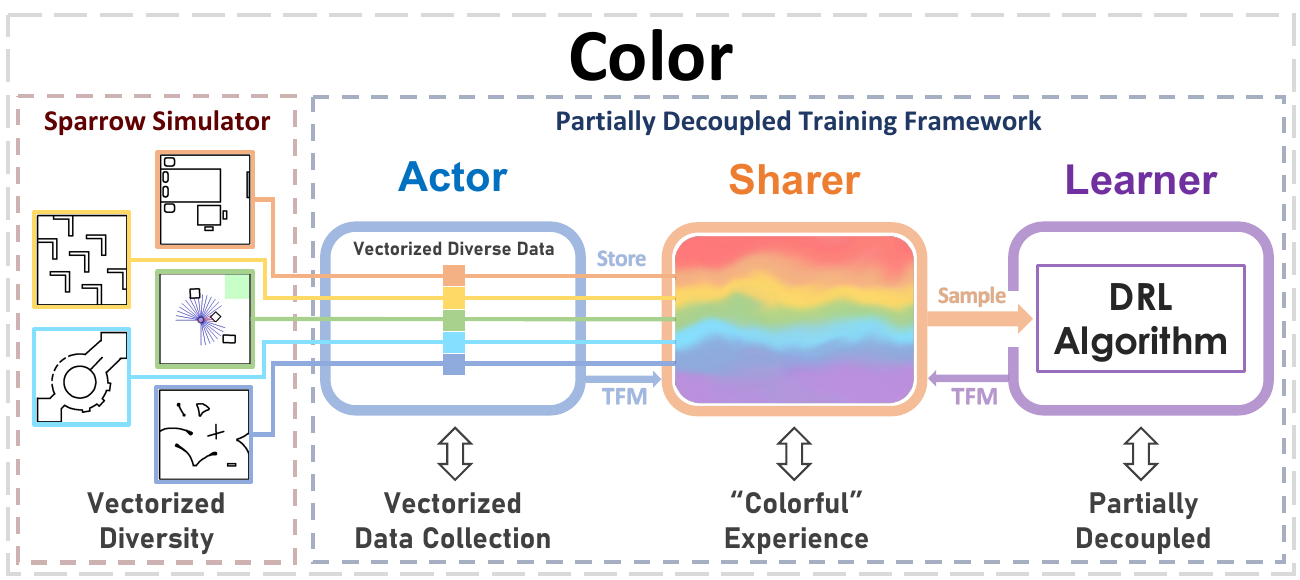}
	\caption{An overview of Color. The left part is the mobile robot-oriented simulator Sparrow, where simulation parameters (such as control interval, control delay, fraction, inertia, velocity range, and sensor noise) and training maps can be readily diversified via vectorized environments. The right part illustrates our efficient DRL training framework, ASL. The seamless integration between Sparrow and ASL is achieved through their interdependent vectorized environments. Notably, Color exhibits the capability to rapidly train a DRL-based local path planner with high generalization capacity.}
	\label{Color}
\end{figure}

\begin{table}[H]
\footnotesize
\center
\caption{\textcolor{black}{Nomenclature}}
\label{tab:Nomenclature}%
\begin{tabular}{lp{6cm}lp{6cm}}
		\hline
\textbf{Symbols} & \textbf{Definition} & \textbf{Abbreviation} & \textbf{Definition} \\
		\hline
    $T$   & Maximal interaction steps & DRL   & Deep Reinforcement Learning \\
    $s$   & State & LPP   & Local Path Planning \\
    $s'$  & Next state after $s$ & RAM   & Random Access Memory \\
    $a$   & Action & RTF   & Real-Time Factor \\
    $\hat{a}$ & VEM processed action & VEM   & Vectorized $\epsilon$-greedy Exploration Mechanism \\
    $r$   & Reward & DR    & Domain Randomization \\
    $done$ & Termination signal of environments & Sim2Real & Simulation to Real world \\
    $\pi$ & Policy & Task2Task & Task to Task \\
    $C$   & Learning start steps & MDA mode  & Multi-thread Deployed Actor mode \\
    $U$   & Model upload frequency & GPU   & Graphics Processing Unit \\
    $\epsilon$ & Exploration noise of $\epsilon$-greedy & CPU   & Central Processing Unit \\
    $E_{min}$ & Minimal exploration noise of VEM & TFM   & Time Feedback Mechanism \\
    $E_{max}$ & Maximal exploration noise of VEM & 2D    & 2-Dimensional \\
    $B$   & Batch Size & 3D    & 3-Dimensional \\
    $B_{step}$ & Total backpropagation steps & SEC & Sample Efficiency Comparison \\
    $T_{step}$ & Total interacting steps & TTEC & Training Time Efficiency Comparison \\
    $V_{step}$ & Total vectorized interacting steps & FP    & Forward Propagation \\
    $N$   & Number of vectorized environments & OI    & ExplOIting interval \\
    $B^{T}_{step}$ & The period of a model optimization procedure & OR    & ExplORing interval \\
    $V^{T}_{step}$ & The period of a VDC procedure & VDC   & Vectorized Data Collection \\
    $\rho$ & Denote $(N \times TPS)/B$ & TPS   & Transitions Per Step \\
    $\tau$ & The elapsed time of ASL & ASL   & Actor-Sharer-Learner \\
    $\gamma$ & Discount factor in MDP & MDP   & Markov Decision Process \\

    $\xi$ & Denote $\rho B^{T}_{step}- V^{T}_{step}$ & GPP & Global Path Planning \\
    $\theta_{a},\theta_{s},\theta_{l}$ & Policy model parameters saved in Actor, Sharer, and Learner & RL-API & Reinforcement Learning Application Programming Interfaces \\
    $v^{t}_{l}, v^{t}_{a}$ & Linear, angular velocity of the robot in Sparrow at timestep $t$ & CFDF  & Conversion-Free Data Flow \\
    $\hat{v}^{t}_{l}, \hat{v}^{t}_{a}$ & Target linear, angular velocity of the robot in Sparrow at timestep $t$ & KB    & KiloByte \\

    $K$   & Kinematic coefficient in Sparrow & MB    & MegaByte\\
    $x,y$ & Robot's coordinates in Sparrow & RAR & Real maps Arrival Rate  \\
    $\phi, \alpha$ & Robot's absolute and relative orientation in Sparrow & TAR & Test maps Arrival Rate \\
    $\Delta t$ & Temporal simulation interval in Sparrow & &  \\
    $O_{xy}$   & Coordinates of obstacle-occupied grids in Sparrow & & \\
    $L$   & Training map size in Sparrow & &  \\
    $D$   & Maximum local planning distance & &  \\
		\hline
	\end{tabular}%
\end{table}%

\section{Introduction}
\textcolor{black}{Local Path Planning (LPP)} portrays a pivotal role in the autonomous navigation of mobile robots, endowing them with the capability to execute unmanned missions such as disaster rescue, military reconnaissance, and material distribution. Given the static path generated by Global Path Planning (GPP) algorithms, the LPP yields a collision-free path in accordance with certain metrics to circumvent the newly arisen obstacles while navigating. Traditional LPP approaches, such as Dynamic Window Approach \citep{DWA} and Fuzzy Logic Algorithm \citep{FLA}, predominantly rely on expert knowledge to deliberately engineer their parameters. Nonetheless, these handcrafted fixed parameters can incur performance deterioration when confronted with complex environments \citep{DWA_flaw,UAV_review}.

\textcolor{black}{Deep Reinforcement Learning (DRL)} is promising to alleviate the aforementioned issue. In DRL, the agent learns by interacting with the environment and is capable of attaining human-level decision-making and control performance. To date, DRL has made remarkable strides in fields such as video games \citep{DQN}, chess \citep{AlphaGo}, robotics \citep{EAAI}, and ChatGPT \citep{ChatGPT}. Nevertheless, despite these momentous breakthroughs, the applications of DRL have been predominantly restricted to the virtual or simulated world. The disappointing efficiency and generalization of DRL have formed a longstanding impediment to its real-world applications: 
\mbox{(1) \textit{Sample efficiency}}: the training of the DRL agent is exceptionally demanding in terms of training samples. \textcolor{black}{This challenge arises primarily from the low quality of the collected training data, as typically, only a small portion of the collected samples significantly contributes to effective learning \citep{PER}.}
(2) \textit{Training time efficiency}: it could take days or even weeks to train a desirable DRL agent in complex environments. \textcolor{black}{This prolonged duration predominantly stems from the slow pace of sample generation, collection, and consumption inherent in DRL training \citep{apex}. }
(3) \textit{Simulation to Real world generalization (Sim2Real)}: to sidestep the unacceptable training time in the real world, researchers typically resort to accelerated physics simulation platforms, with the expectation that the model trained in a simulator can be transferred to real-world situations. \textcolor{black}{However, the physical discrepancy between the simulated and the real world is challenging to diminish, thus significantly impeding the transfer \citep{DR}.}
(4) \textit{Task to Task generalization (Task2Task)}: the DRL agent is susceptible to overfitting in the training environments, leading to performance degradation when exposed to new or altered tasks. \textcolor{black}{Such flaw is mostly due to the insufficient diversity of training environments \citep{ppo_dm}.}

\textcolor{black}{
To address these challenges, a solution termed Color is proposed, featuring a highly efficient training framework, ASL, alongside a fast and diverse simulator named Sparrow. First, we approach the challenge of sample efficiency via enhancing the quality of the collected training samples, achieved through ASL’s vectorized exploration mechanism. In addition, Sparrow is also equipped with diverse simulation abilities to enable the generation of high-quality training samples. Second, we reduce training time by accelerating data generation, collection, and consumption. Specifically, Sparrow’s vectorized (parallel) simulation significantly contributes to rapid data generation, while ASL's vectorized data collection mode effectively expedites data collection. For data consumption, ASL’s partially decoupled design allows for simultaneous data collection and consumption, preventing the suspension of model optimization during environmental interactions. Third, we narrow the discrepancy between simulated and real world via incorporating Domain Randomization \citep[DR,][]{DR} into the vectorized simulation of Sparrow, aiming to enhance the Sim2Real capacity. Finally, we tackle the Task2Task issue by Sparrow’s vectorized diversity. This feature enables parallel simulation of diverse maps, exposing the trained agent to a wide range of tasks and thereby improving its generalization capability across different tasks.} In summary, the main contributions of this paper can be outlined as follows:\color{black}
\begin{enumerate}[1)]
\item A partially decoupled training framework \textcolor{black}{with vectorized data collection}, namely Actor-Sharer-Learner (ASL), is proposed to improve the training efficiency of DRL algorithms. Experiments demonstrate that off-policy DRL algorithms with experience replay can be readily integrated into our framework to enjoy a remarkable promotion.

\item A lightweight mobile robot-oriented simulator named Sparrow has been developed. The lightness of Sparrow boosts its vectorization, which not only expedites data generation but also enables vectorized diversity, allowing diverse simulation setups across different copies of the vectorized environments to enhance the generalization of the trained DRL algorithms.

\item Combining the ASL with Sparrow through their interdependent vectorized environments, we formulate our DRL solution to the LPP problem, referred to as Color. Through one hour of simulation training, the Color is capable of yielding a real-world local path planner with laudable resilience to generalize over a wide variety of scenarios.
\end{enumerate}

The remainder is organized as follows. Related works concerning DRL's efficiency, generalization, and commonly used simulators are reviewed in Section \ref{section2}. The ASL training framework and the Sparrow simulator are introduced in Section \ref{section3}. Comprehensive experiments involving the evaluation of ASL, Sparrow, and the trained local path planner, as well as ablation studies, have been conducted in Section \ref{section4}. The conclusion is derived in Section \ref{section5}.

\section{Related works}
\label{section2}

This section reviews related works that aim to enhance the efficiency and generalizability of DRL algorithms, as well as the simulation platforms commonly utilized in DRL. 
Each subsection concludes with an analysis of the limitations of these works.

\color{black}
\subsection{Efficiency of DRL}
\label{section2.1}
The efficiency of DRL encompasses sample efficiency and training time efficiency. Methodologies for enhancing sample efficiency primarily focus on the following four perspectives:
(1)	Sample expansion: Expanding the training samples via data transformation or augmentation techniques, such as TLDA \citep{TLDA} and CoIT \citep{coit}.
(2)	Distributional Q-value representation: Representing Q-values as probability distributions instead of point expectation estimates, as seen in methods like C51 \citep{c51}, DSAC \citep{DSAC-v1}, and DSAC-T \citep{DSAC-T}.
(3)	Neural network architecture optimization: Designing neural network architectures optimized for DRL, with notable examples including Dueling DQN \citep{duel} and Noisy Nets \citep{noisy}.
(4)	Prioritized training strategies: Utilizing training samples in a prioritized manner, exemplified by PER \citep{PER}, QER \citep{QER}, and PHER \citep{realworld_human}.
A common limitation of the aforementioned methodologies is the suspension of the training procedure during the agent's interactions with the environment. This alternation between model optimization and data collection phases results in suboptimal training time efficiency.

In the context of enhancing training time efficiency, the DRL community generally resorts to decoupled training frameworks. 
\citet{A3C} proposed the A3C framework, which decouples the data collection from model optimization, significantly reducing training time compared to standard DRL methods. The IMPALA \citep{impala} framework enhances A3C with off-policy correction and multiple synchronous optimizations, rendering it more efficient for scaling to large learning systems. However, a shared limitation of both A3C and IMPALA is their low sample efficiency, as the training data is discarded after a single use.
To mitigate this limitation, experience replay has been integrated into decoupled frameworks to facilitate sample reuse, as exemplified by the seminal Ape-X framework \citep{apex}. 
\color{black}
Specifically, Ape-X decouples data collection and model optimization into multi-actor and single-learner components, which operate concurrently to enhance training time efficiency. Unlike previous frameworks, Ape-X stores the data collected by actors in an experience buffer for reuse, rather than discarding it after use. Similar works following this framework include R2D2 \citep{r2d2}, SEED RL \citep{seed_rl}, and QR-SAC \citep{qr_sac_nature, QR-SAC}.
Nevertheless, two notable drawbacks persist within this framework. First, the Multi-thread Deployed Actor (MDA) mode is inefficient, which could exert three adverse repercussions upon the learning system: a) if the Forward Propagation (FP) of the network is computed by the CPU, it precludes the usage of large-scale networks because the CPU is notorious for its inefficiency in large scale networks computing. b) if the FP is computed by the GPU, the parallel computing feature of the GPU is not fully utilized, for the computation is requested separately and asynchronously by many actors. c) the data flow can be cumbersome and bottlenecked by the transmission bandwidth, where the data has to be firstly accumulated by the local buffer of each actor and then packed into the shared memory. 
Second, the actors and learner are completely decoupled without coordinating their running speeds. If actors run far slower than the learner, it’s likely that the data is overused, which could adversely impact its stability. Otherwise, the data might be abandoned without full utilization, resulting in low sample efficiency. Since the relative speed of the actors and learner can be influenced by the breakdown of actors or fluctuates with the hardware occupation \citep{apex}, these two unfavorable cases are inevitable.
\color{black}

In summary, methodologies aimed at improving sample efficiency are often constrained by low training time efficiency, while those focused on enhancing training time efficiency face challenges related to MDA mode inefficiency and issues of sample overuse or underuse within decoupled frameworks. Therefore, further research is essential to overcome these limitations and close the gaps in current studies.

\subsection{Generalization of DRL}
The generalization of DRL can be bifurcated into Sim2Real and Task2Task generalization. Methods concerning improving the Sim2Real generalization primarily involve three categories: 
(1) System identification: Define and tune the parameters to match the real-world environment. \citet{realworld_nature} utilized a residual model to quantify the gap between simulated and real-world environments, wherein Gaussian processes and k-nearest-neighbor regression were applied to model discrepancies in perception and dynamics, respectively. After pre-training in a conventional simulator, the agent was fine-tuned in a residual model-enhanced simulation to facilitate Sim2Real transfer. 
(2) Domain randomization: Randomize the simulation setups while training, such that the real-world environment can be deemed as another variation of these randomized environments. \citet{realworld_30hour} developed a DR-augmented simulation platform DMHouse, where the layout of the indoor room, along with the lighting conditions, were randomized during training to ensure the generalization across diverse real-world conditions. 
(3) Domain adaptation: Adapt the trained model by learning mapping or invariant features from simulation to the real world, or re-training in the real world. \citet{realworld_human} proposed a human-guided reinforcement learning framework, allowing humans to guide the robot in the real world, wherein the DRL agent can be re-trained by learning from real-world demonstrations. In doing so, the fine-tuned agent can be adaptable to more corner cases from the real world.

Regarding Task2Task generalization, methodologies can be broadly categorized into three main classifications:
(1) Training regularization. Notable methods include injecting tangent prop into the critic loss function \citep{tprop}, averaging Q target and Q function over transformed input images \citep{DrQ}, and incorporating L2-regularization, dropout, and batch normalization into DRL training \citep{CoinRun}.
(2) Training environment diversification. For instance, \citet{LiZhi} randomly initialized the agent and environment at the start of each training episode to enhance diversity, while \citet{Procgen} randomly varied the visual properties of the environments to achieve a similar effect.
(3) Training data augmentation. \citet{TLDA} proposed a task-aware data augmentation technique to identify and augment the task-correlated part of the input image. \citet{RandomConv} utilized random convolution networks to augment training images within the feature space. \citet{ADA} developed an automatic data augmentation diagram to autonomously discover effective augmentation tools for DRL tasks.

Although numerous studies have explored the generalization of DRL, their practicality remains debated, as some approaches necessitate expert knowledge \citep{realworld_nature} of specific tasks or human assistance \citep{realworld_human} during training. 
Moreover, prior research has primarily focused on the standard DRL framework, without considering the integration of these techniques with high-efficiency training frameworks, such as the decoupled framework. Therefore, further investigation into the generalization of DRL is warranted and could be a desirable incentive to the wide-ranging application of DRL.

\subsection{Simulation platform}
\label{section2.3}

Based on the underlying calculation, DRL simulation platforms can be categorized into (i) non-physics engine-based and (ii) physics engine-based. Gym \citep{gym}, CoinRun\citep{CoinRun}, and Procgen\citep{Procgen} are three widely utilized non-physics engine-based platforms. Gym is a fundamental toolkit for evaluating and comparing DRL algorithms, providing a variety of environments such as classic control problems and video games. This makes it suitable for both beginners and researchers seeking rapid prototyping and algorithm testing. CoinRun is a video game environment where agents must navigate obstacles and collect coins to achieve their goals. The procedurally generated visual diversity of CoinRun provides a robust testbed for assessing the robustness and adaptability of DRL algorithms, particularly in evaluating their ability to non-seen tasks. Procgen extends the CoinRun environment by providing a suite of 16 distinct procedurally generated games, offering a more comprehensive platform for quantifying the generalization of DRL algorithms. 

Regarding physics engine-based platforms, notable examples include Robotics Gym \citep{robogym}, DeepMind Control Suite \citep{dm_control}, PyBullet \citep{PyBullet}, ML-Agents \citep{ML_Agent}, Webots \citep{Webots}, and Gazebo \citep{gazebo}. 
Robotics Gym and DeepMind Control Suite are two platforms built upon the MuJoCo physics engine, developed by OpenAI and Google, respectively. Both platforms provide a series of continuous control tasks, making them valuable tools for robotic control research. PyBullet is a high-fidelity physics simulation library that offers a Python interface for the Bullet physics engine, commonly used in DRL research for scenarios requiring precise physical interactions, such as robotic manipulation, locomotion, and collision detection. ML-Agents is an open-source platform powered by the Unity game engine, characterized by its integration with advanced DRL algorithms to facilitate rapid deployment in 2D, 3D, and VR/AR games, thereby serving as a convenient tool for both researchers and game developers. 
\color{black}
Webots, developed by Cyberbotics, is a robot simulation software distinguished by its user-friendly interface and integration with multiple programming languages. It supports a wide range of robotic platforms and sensors and provides an extensive suite of tools for modeling, simulating, and analyzing robot behavior in various environments, enabling researchers to explore different configurations and scenarios within simulation. The Gazebo robotics simulator, developed by Open Robotics, is an open-source platform that facilitates the design, development, and testing of robotic systems. Analogous to Webots, Gazebo offers a comprehensive suite of tools and libraries for the creation of realistic simulation environments. Its high-fidelity physics engine, advanced visualization capabilities, and versatile sensor and robotics simulation functionalities enable users to construct virtual settings that closely mirror real-world conditions. Notably, Gazebo's seamless integration with the Robot Operating System (ROS) \citep{ros} allows researchers to develop and evaluate robotic systems in a virtual setting, with the ability to then rapidly deploy these systems to real-world scenarios, rendering it a popular choice for industrial applications. 

Despite advancements, it is important to note that non-physics engine-based platforms are specifically designed for virtual environments, lacking support for realistic robotics simulation. This limitation restricts their applicability to agents facing real-world scenarios.
Conversely, physics engine-based platforms are hindered by inherently high computational costs, primarily due to the complex physics calculations necessary to preserve the fidelity and realism of simulated environments. These demands can result in suboptimal simulation speeds \citep{Gazebo_slow2,PyBullet_slow,Gazebo_slow3,CCPPO}, which can be further exacerbated by the inefficiencies of DRL algorithms, resulting in excessively prolonged training time. 
Therefore, further research is needed to develop more advanced simulation platforms for DRL, promoting efficient training and effective generalization.
\color{black}

\section{Methodology}
\label{section3}

In this section, we will first elaborate on our efficient training framework ASL and our mobile robot-oriented simulator Sparrow, and then discuss how to elegantly combine them together, fully taking their respective advantage, to form the DRL solution to LPP problems, namely the Color.

\subsection{Actor-Sharer-Learner training framework}
Section \ref{section2.1} reveals a contradiction between sample efficiency and training time efficiency. Despite extensive efforts to tackle these challenges, existing approaches remain separate. We argue that both the sample and training time efficiency are prominent for DRL, and improving them simultaneously could considerably boost the application of DRL. In pursuit of this goal, we proposed the ASL training framework. 

\textcolor{black}{To equip our framework with high training time efficiency, we inherit the decoupled architecture from Ape-X. As shown in Fig. \ref{ASL}, an actor and a learner are employed to collect data and optimize the model individually. The actor and learner run concurrently in different threads, and the information that needs to be shared is managed by a sharer. As aforementioned, the Ape-X is sorely restricted by its MDA mode and completely decoupled architecture. We tackle the first issue by resorting to the vectorized environments \citep{envpool} and the second issue by adding a Time Feedback Mechanism (TFM) to softly connect the actor and the learner. }

\subsubsection{Vectorized data collection}
The vectorized environments run multiple copies of the same environment parallelly and independently, and the Vectorized Data Collection (VDC) mode of our ASL framework interacts with these environments in a batched fashion. The superiority of VDC is evident. First, the parallel computation advantage of GPU could be exploited to the fullest extent thanks to the batched interaction. Second, when interacting with vectorized environments of $N$ copies, the VDC mode necessitates only a single instantiation of the actor network, leading to a substantial reduction in hardware resource requirements in contrast to the multi-thread deployed actor mode that requires $N$ instantiations. Third, the batched data is stored as a whole in the experience buffer, eliminating the need for the thread lock stemming from asynchronous data preservation requests caused by multiple actors. Accordingly, it is reasonable to imagine that the three merits could additionally shorten the time needed for training.

\begin{figure}[t]
	\centering
	\includegraphics[width=0.56\textwidth]{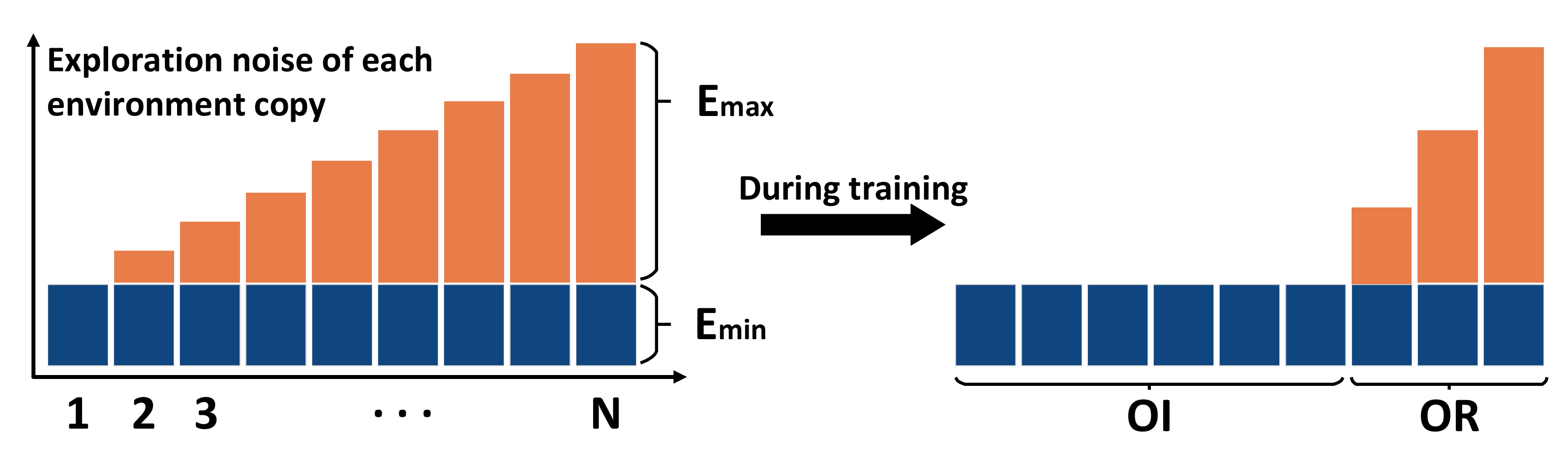}
	\caption{Vectorized $\epsilon$-greedy Exploration Mechanism. Here, the horizontal axis is the index of the vectorized environments, and $N$ is the total number of vectorized environments.}
	\label{VEM}
\end{figure}

\begin{figure}[t]
	\centering
	\includegraphics[width=0.72\textwidth]{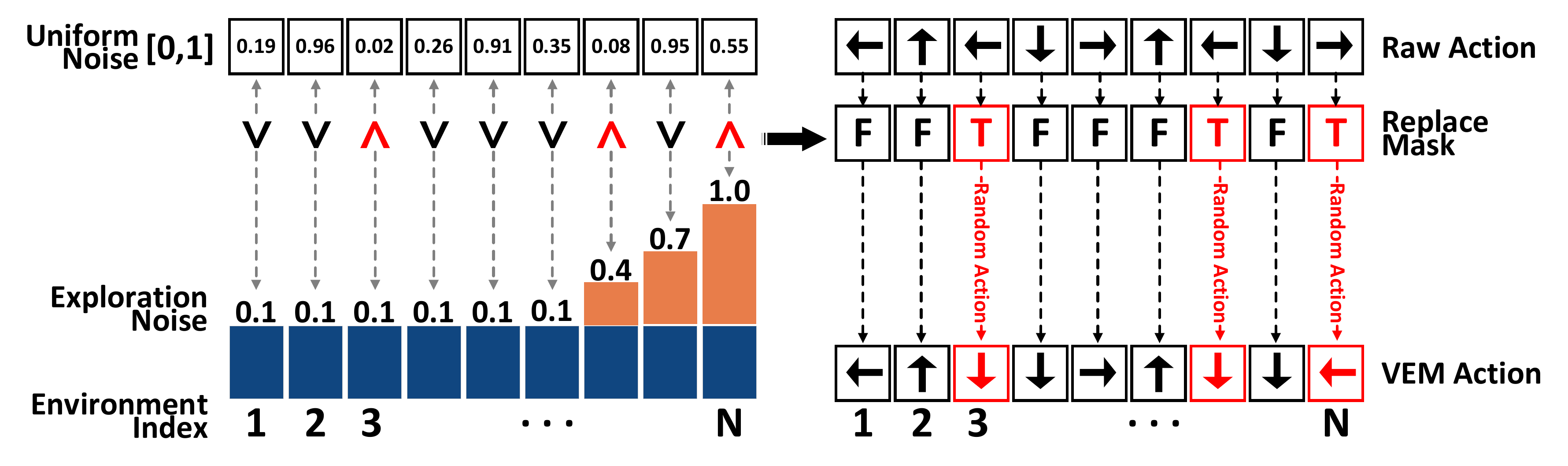}
	\caption{\textcolor{black}{VEM-based exploration. Here, the raw action is generated by the DRL model, and the VEM action will be applied to the N distinct copies of vectorized environments.}}
	\label{VEM_replace}
\end{figure}

Correspondingly, a Vectorized $\epsilon$-greedy Exploration Mechanism (VEM) is devised. In canonical $\epsilon$-greedy exploration \citep{sutton2018}, the agent executes its own policy at a probability of $1-\epsilon$ and explores the environment at a probability of $\epsilon$. The exploration noise $\epsilon$ is linearly or exponentially decreasing during training. Although the $\epsilon$-greedy exploration is promising to maintain a balance between exploitation and exploration, its limitation is self-evident. Exploration is only guaranteed at the beginning of training and is eventually replaced with exploitation due to the decrease in noise. We argue this setting can be irrational in some contexts, especially when the environment is stage-by-stage. For instance, in some video games, the agent takes extensive exploration to pass the current stage, only to find the exploration noise is decreased to a low level, and sufficient exploration can no longer be guaranteed in the new stage. Contrastingly, the VEM divides the vectorized environments into expl\textbf{oi}ting interval (OI) and expl\textbf{or}ing interval (OR). As shown in Fig. \ref{VEM}, the exploration noises maintain the minimal value $E_{min}$ in OI and are linearly interpolated from $E_{min}$ to $E_{min}+E_{max}$ in OR. 
\textcolor{black}{Every copy of the vectorized environments is allocated with their respective exploration noise, ranging from $E_{min}$ for the 1st copy to $E_{min}+ E_{max}$ for the Nth copy.  At every interaction, the raw action generated by the DRL agent is substituted with a random action based on the corresponding exploration noise, as illustrated in Fig. \ref{VEM_replace}.}
In addition, the OR is linearly diminished to a certain extent during the training process. Relative to standard $\epsilon$-greedy exploration, the VEM inherits its positive attribute: progressively focus on exploitation during training. Meanwhile, the VEM also bypasses its flaw: conserve a sound exploration capability even in the final training stage. Consequently, the VEM reaches a more sensible balance between exploration and exploitation, thus enhancing the sample efficiency accordingly.
\textcolor{black}{It should be noted that the VEM incorporates four hyperparameters: $E_{min}$, $E_{max}$, final OR, and the decay steps of OR. These hyperparameters necessitate environment-specific tuning. Elevating these hyperparameters results in VEM emphasizing exploration and vice versa. }

\subsubsection{Time feedback mechanism}

In off-policy DRL with an experience replay mechanism, one can train the model multiple times after one interaction with the environment or a single update after multiple interactions. To quantify the data utilization rate, we define the \textit{Transitions Per Step (TPS)}:

\begin{equation}
	\label{TPS}
	\mbox{\textit{TPS}} = \frac{B \times B_{step}}{T_{step}}
\end{equation}

\begin{figure*}[!t]
	\centering
	\includegraphics[width=0.9\textwidth]{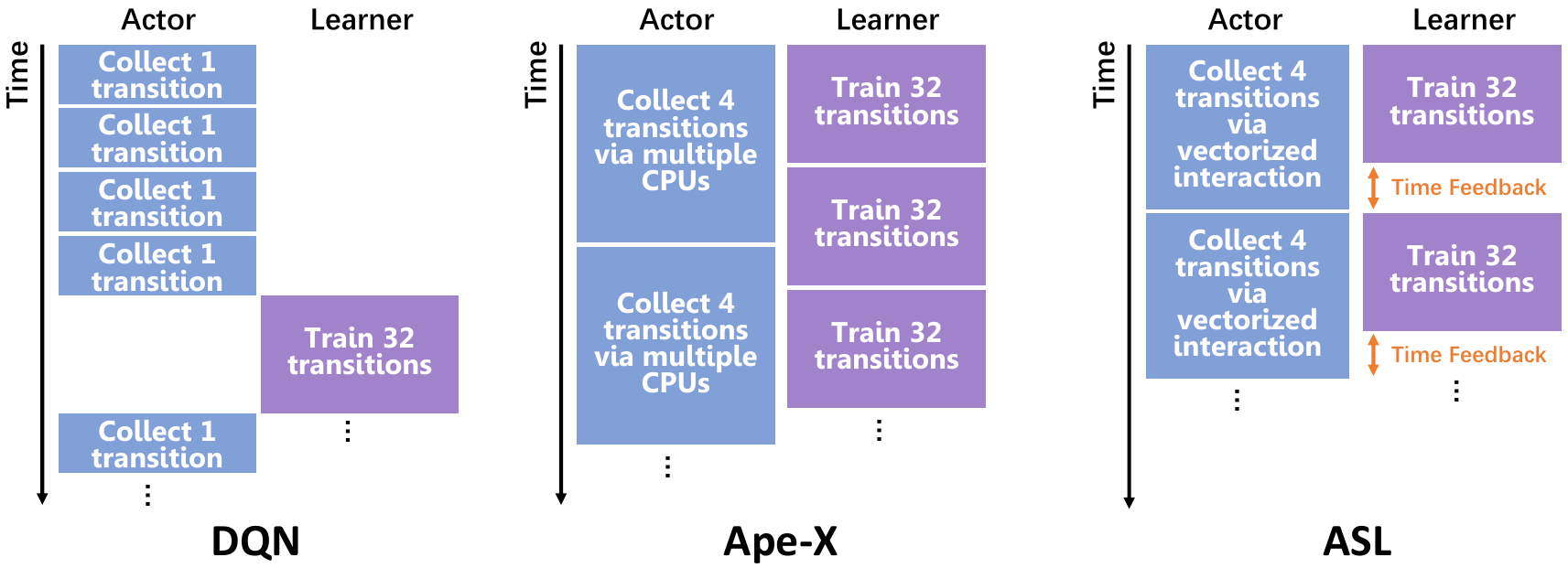}
	\caption{Illustration of the coupled framework (left, representative: DQN), completely decoupled framework (middle, representative: Ape-X), and partially decoupled framework with TFM (right, representative: ASL). In this example, both the DQN and the ASL seek to maintain a $\mbox{\textit{TPS}}$ of 8. However, the cycle (the collection of 4 transitions + the training of 32 transitions) in DQN is more time-consuming due to its step-by-step collection and the alternation between collection and training. Conversely, the ASL demonstrates computational efficiency attributable to its vectorized interaction and its ability to perform collection and training concurrently. In contrast, the Ape-X fully decouples the collection and training processes, resulting in an uncontrolled $\mbox{\textit{TPS}}$ and risking its training stability and sample efficiency.}
	\label{TFM}
\end{figure*}

\noindent where $B$ is the batch size, $B_{step}$ is the total backpropagation steps (the number of times that the model has been optimized), and $T_{step}$ is the total interacting steps (the number of transitions has been collected so far). The \mbox{\textit{TPS}} indicates how many transitions have been replayed for training every after a new transition has been collected. A large \mbox{\textit{TPS}} implies a high data utilization rate but may also lead to model collapse due to data overuse \citep{overuse}. Conversely, a small \mbox{\textit{TPS}} results in data underuse, resulting in low sample efficiency. Since the Ape-X expedites the training at the expense of completely decoupling the interaction and optimization procedures without controlling the \mbox{\textit{TPS}}, it is inevitable to incur a loss of stability or sample efficiency. In contrast, the ASL employs the TFM to modulate the relative speed between actor and learner, imposing a certain amount of dormant period upon the faster one to coordinate the \mbox{\textit{TPS}} of the learning system, as illustrated by Fig. \ref{TFM}. In doing so, the data overuse or underuse phenomenon could be circumvented.
\color{black}
In vectorized environments, based on (\ref{TPS}), we have the following equation:
\begin{equation}
	\label{V2B}
	B_{step} = \frac{N \times \mbox{\textit{TPS}}}{B} V_{step}
\end{equation}

\noindent where $N$ is the number of copies of vectorized environments and $V_{step} = T_{step}/N$ is the total number of vectorized steps. Eq. (\ref{V2B}) implies that to maintain a stationary $\mbox{\textit{TPS}}$,  the ratio of $B_{step}$ and $V_{step}$ should be consistently held at $(N \times \mbox{\textit{TPS}})/B$. 
To connect (\ref{V2B}) with time feedback, we define the period, namely the elapsed time, of one VDC procedure and one model optimization procedure as $V_{step}^{T}$ and $B_{step}^{T}$, respectively. Meanwhile, we denote the elapsed time of ASL as $\tau$. Due to the fact that the actor and learner run concurrently, the elapsed time of ASL is equivalent to that of the actor and the learner. Thus, $B_{step}=\tau / B_{step}^{T}$ and $V_{step}=\tau / V_{step}^{T}$. Then, we have
\begin{equation}
	\label{Times2Period}
	\frac{\tau}{B_{step}^{T}}  = \rho \frac{\tau}{V_{step}^{T}}
\end{equation}
\noindent where $\rho=(N \times \mbox{\textit{TPS}})/B$. Evidently, based on Eq. (\ref{Times2Period}), the following equation holds:
\begin{equation}
	\label{TB2TV}
	V_{step}^{T}  = \rho B_{step}^{T} 
\end{equation}

\color{black}
Eq. (\ref{TB2TV}) suggests that by modulating the two elapsed times in a manner consistent with such relative relationship, a fixed \mbox{\textit{TPS}} can be approximately upheld. To achieve this end, the actor and learner would record their individual period $V_{step}^{T}$ and $B_{step}^{T}$. The sharer then coordinates their running speed according to the two following principles:

\begin{itemize}
	\item{if $\xi = \rho B_{step}^{T} - V_{step}^{T} > 0$, the actor sleeps for a period of $\xi$ every after one VDC procedure.}
	
	\item{if $\xi \leq 0 $, the learner sleeps for a period of $ -\xi/{\rho} $ every after one model optimization procedure.}
\end{itemize}

\textcolor{black}{Essentially, the TFM is a blocking synchronization mechanism that partially (or softly) connects the Actor and Learner to align their data collection/consumption rates. This approach overcomes the respective limitations of DQN and Ape-X: (i) concurrent Actor and Learner operation enhances time efficiency compared to DQN, and (ii) blocking synchronization renders ASL more stable and sample-efficient than Ape-X. Consequently, our TFM maintains a fixed data utilization rate with reduced temporal costs, enhancing the stability and speed of our ASL framework.} Regarding the dormancy in TFM, one might wonder whether the TFM maintains the sample efficiency at the cost of sacrificing the training time efficiency, which deviates from our rudimentary purpose. Indeed, the dormancy could slow down the training framework to some extent but bring about a more sensible data recency, which tremendously facilitates model optimization. That is, reaching the same performance with fewer times of optimization, which conversely reduces the time needed for training.

\subsubsection{Workflow of ASL}

\begin{figure}[H]
	\centering
	\includegraphics[width=0.9\textwidth]{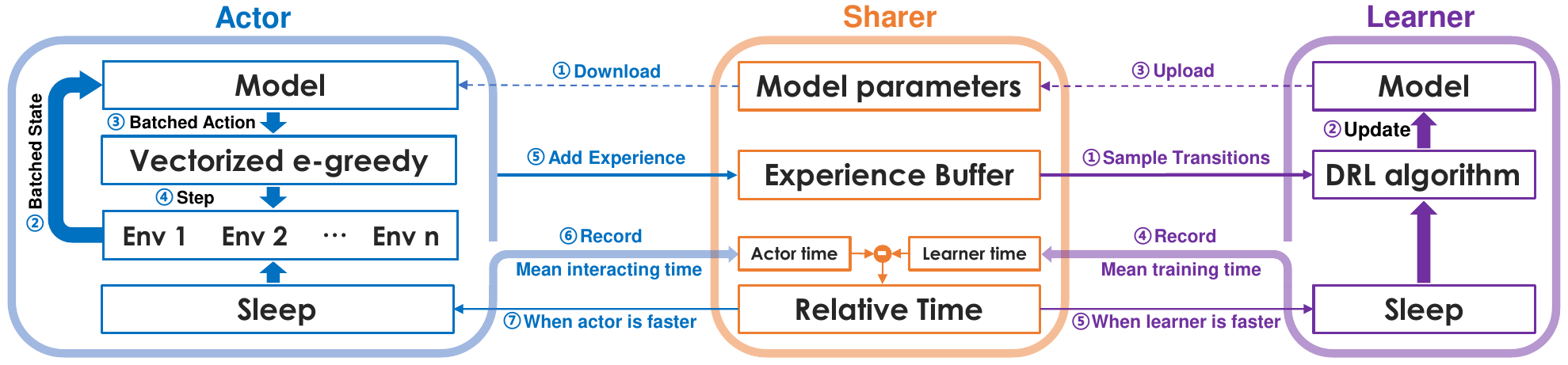}
	\caption{Schematic of the Actor-Sharer-Learner training framework}
	\label{ASL}
\end{figure}

\color{black}
The pseudocode of ASL is given in Algorithm \ref{algo:actor} and \ref{algo:learner}, and the corresponding schematic is shown in Fig. \ref{ASL}. 
To maximize interaction speed, the actor interacts with the vectorized environments in a batched fashion through the VDC mode, where all the batched data is represented by \textit{torch.tensor}\footnote{https://pytorch.org/docs/stable/tensors.html} to leverage GPU acceleration. At each interaction, the process unfolds as follows: the actor receives the batched state $s$ from the vectorized environments, which is then input into the actor model $\theta_a$. This model outputs the raw batched action $a$, which is subsequently replaced with random actions via VEM, yielding batched $\hat{a}$. Finally, the batched $\hat{a}$ is applied to the N distinct instances of vectorized environments.
\color{black}

\begin{algorithm}[H]
\footnotesize
\caption{Actor}\label{algo:actor}
\tcc{
$T_{step}$:total interacting steps;$T$:maximal interaction steps;$N$:number of vectorized environments(envs);
$(s,a,r,s',done)$:(state, action, reward, next state, termination signal); 
$\hat{a}$:VEM processed action;
$\theta_a$:model parameters of Actor; 
$\theta_s$:model parameters of Sharer; 
$\pi(\theta_a)$:policy networks;
$V_{step}^{T}$:the period of a VDC procedure;
$\xi$:the relative time between Actor and Learner
}
$\textbf{s}$ = envs.reset() \tcp{Generate the initial batched state.}
\While{Sharer.$T_{step}<T$}{
$\textbf{1.}$ $\theta_{a} \leftarrow \theta_{s}$ \tcp{Inquire the Sharer whether there is a new model to download.}

$\textbf{2.}$ $\textbf{a}$ = $\pi(\textbf{s}; \theta_{a})$ \tcp{Map the batched state to batched action with the latest policy $\pi(\theta_{a})$.}

$\textbf{3.}$ $\hat{\textbf{a}}$ = VEM$(\textbf{a})$ \tcp{Inject the batched action with stochastic action by the VEM.}
 
$\textbf{4.}$ $ \textbf{s}', \textbf{r}, \textbf{done}$ = envs.step($\hat{\textbf{a}}$) \tcp{Interact with the vectorized envs with VEM processed action $\hat{\textbf{a}}$.}

$\textbf{5.}$ Sharer.buffer.add($ \textbf{s}, \hat{\textbf{a}}, \textbf{r}, \textbf{s}', \textbf{done}$) \tcp{Add the batched transitions to  Sharer.}

$\textbf{6.}$ $V_{step}^{T} \rightarrow$ Sharer \tcp{Record the mean interacting time from Step 1 to 5, and send it to the Sharer.}

$\textbf{7.}$ \textbf{if}  Sharer.$\xi>0$: sleep($\xi$) \tcp{Fetch the relative time $\xi$ from the Sharer, and sleep for $\xi$ seconds if the Actor runs faster.}

$\textbf{8.}$  $ \textbf{s} = \textbf{s}'$, Sharer.$T_{step}$+=$N$, Sharer.buffer.size+=$N$
}\end{algorithm}

\begin{algorithm}[H]
\footnotesize
\caption{Learner}\label{algo:learner}
\tcc{
$T_{step}$:total interacting steps;
$T$:maximal interaction steps;
$C$:learning start steps;
$\rho$:$(N \times \mbox{\textit{TPS}})/B$;
$B_{step}^{T}$:the period of a model optimization procedure;
$\xi$:the relative time between Actor and Learner;
$\theta_l$:model parameters of Learner; 
$\theta_s$:model parameters of Sharer; 
$B_{step}$:total backpropagation steps;
$(s,\hat{a},r,s',done)$:a mini-batch of transitions;
$U$:model upload frequency; 
$DRL(\cdot)$:the underlying Deep Reinforcement Learning algorithm
}
\While{Sharer.$T_{step}<T$}{
\uIf{Sharer.buffer.size $>$ $C$}{
$\textbf{1.}$ $ \textbf{s}, \hat{\textbf{a}}, \textbf{r}, \textbf{s}', \textbf{done}$ = Sharer.buffer.sample() \tcp{Sample a mini-batch of transitions from the Sharer. }

$\textbf{2.}$ $\theta_{l} $ = DRL$(\textbf{s}, \hat{\textbf{a}}, \textbf{r}, \textbf{s}', \textbf{done}; \theta_{l})$ \tcp{Optimize the model with the underlying DRL algorithm.}

$\textbf{3.}$ \textbf{if} $B_{step}\ \%\ U$\ =\ 0: $\theta_{l} \rightarrow \theta_{s}$ \tcp{Upload the latest model $\theta_{l}$ to the Sharer at a fixed frequency $U$.}

$\textbf{4.}$ $B_{step}^{T} \rightarrow$ Sharer \tcp{Record the mean optimization time from Step 1 to 3, and send it to the Sharer.}

$\textbf{5.}$ \textbf{if}  Sharer.$\xi\leq 0$: sleep($ -\xi/{\rho} $) \tcp{Fetch the relative time $\xi$ from the Sharer, and sleep for $ -\xi/{\rho} $ seconds if the Learner runs faster.}

$\textbf{6.}$ $B_{step}$+=$1$
}}\end{algorithm}

\color{black}

On the other hand, the learner is encapsulated as a separate module and operates simultaneously with the actor until reaching the maximum interaction step $T$. More detailedly, the learner samples data from the sharer, trains the learner model $\theta_l$ with the underlying off-policy DRL algorithm, and uploads the model at a fixed frequency $U$. The relative running speed between the learner and actor is modulated by the TFM to maintain a consistent data utilization rate. Note that, to mitigate overfitting, the learner process is initiated only after a certain number of transitions have been collected.

The sharer manages the data necessitating sharing between actor and learner, encompassing model parameters, experience buffer, and relative runtime. This data sharing is implemented via Python's \textit{BaseManager}\footnote{https://docs.python.org/3/library/multiprocessing.html}, which efficiently manages shared data across multiple processes. To prevent data occupancy conflicts, a data lock mechanism is developed. This lock is operationalized by associating the shared data with an occupancy flag. When either the actor or learner is utilizing the data, the flag is set to \textit{True}, thereby blocking access from the other process until the data becomes available. This approach effectively mitigates scenarios where, for instance, the learner attempts to access data concurrently being modified by the actor.

\color{black}
To sum up, the superiority of our ASL training framework can be summarized as follows:
\begin{itemize}
\item The VDC operates in a batched manner, enabling GPU acceleration, reducing neural network overhead, simplifying data preservation, and ultimately enhancing training time efficiency. 
\item The VEM within the VDC progressively emphasizes exploitation during training while maintaining sound exploration capabilities even in the final training stages. This approach achieves a more reasonable balance between exploration and exploitation, thereby enhancing sample efficiency.
\item The data collection (actor) and model optimization (leaner) processes are partially decoupled and operate concurrently, which saves time from alternation and improves training time efficiency eventually.
\item The actor and learner are partially connected by the TFM to coordinate the data collection/consumption rate, thus circumventing data overuse or underuse phenomenon and elevating sample efficiency and training stability eventually. 
\end{itemize}
\color{black}

\color{black}
\subsection{Sparrow simulator}

\begin{figure}
	\centering
	\subfloat[]{\includegraphics[width=0.4\textwidth]{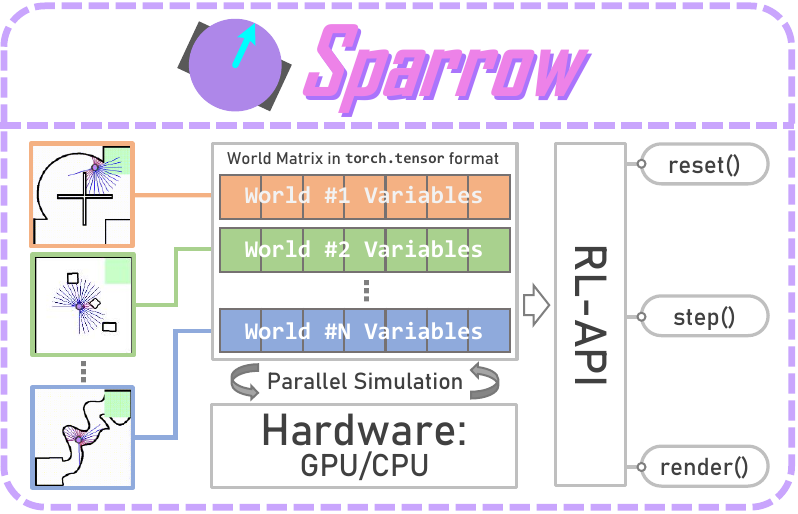}
		\label{Sparrow}}
	\hfil
	\subfloat[]{\includegraphics[width=0.214\textwidth]{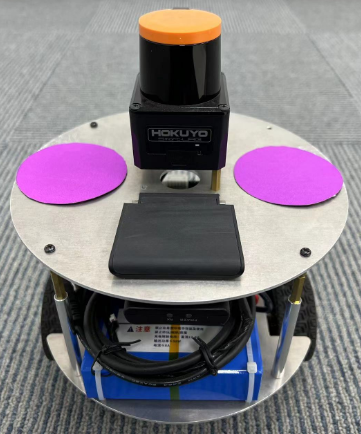}
		\label{robot}}
	\caption{(a) An overview of the Sparrow simulator. In the rendered maps, the green area represents the target area; the black rectangles are the obstacles; the blue rays and the purple circle denote the LiDAR and the robot, respectively. In Sparrow, each world is abstracted as a vector so that the vectorized environments can be represented by a matrix to facilitate parallel simulation. The Sparrow simulator interacts with external algorithms through its RL-API. (b) The real-world counterpart of the robot in Sparrow: a differential wheeled robot.}
	\label{S&R}
\end{figure}

Sparrow is a mobile robot-oriented simulation platform, as shown in Fig. \ref{S&R} (a). The development of Sparrow is triggered by the niche, as detailed in Section \ref{section2.3}, between two types of widely adopted platforms. The physics engine-based platforms are inordinately heavy and thus form an impediment to parallelization as well as diversity, while the non-physics engine-based platforms are merely lightweight virtual environments, unable to foster the real-world application of DRL. Consequently, we posed a question to ourselves: is it possible to propose a mobile robot-oriented simulator that is lightweight to expedite data acquisition and diversifiable to augment the generalization capacity of the trained agent? 

To this end, we developed Sparrow, which utilized 2D occupancy grid maps and simplified kinematics to describe the obstacles and the robot, circumventing the complicated physics calculation and thereby contributing to its lightness. Meanwhile, Sparrow’s world matrix representation facilitates vectorization, thus augmenting simulation speed and data throughput. Notably, due to Sparrow’s lightweight design, the hardware resources would not be extravagantly consumed even with extensive vectorization. Moreover, Sparrow supports domain randomization across different copies of the vectorized environments, termed vectorized diversity. This feature enables the generation of “colorful” training data, thereby enhancing the generalization capability of the trained DRL model. Prior to diving into these specifics, we begin with a concise summary of Sparrow's key functionalities and features:

\begin{itemize}
\item Functionalities
  \begin{itemize}
    \item[$\circ$] Robot motion simulation: efficient motion simulation with simplified kinematics
    \item[$\circ$] LiDAR simulation: range finding with 2D occupancy grid maps and ray casting algorithm \citep{raycast, CCPPO}
    \item[$\circ$] Sensor noise simulation: adjustable noise on simulation results
    \item[$\circ$] Collision detection: collision detection between robot and obstacles
    \item[$\circ$] Map customization: binary image maps, easy to customize with image editing software 
    \item[$\circ$] Vectorization: parallel simulation with world matrix representation, fast simulation speed and high data throughput
    \item[$\circ$] Vectorized diversity: domain randomization across different copies of the vectorized environments, enabling “colorful” training data
    \item[$\circ$] AutoReset: an independent auto-reset mechanism for vectorized environments
  \end{itemize}

\item Features
  \begin{itemize}
    \item[$\circ$] Fast and lightweight: 815.5 real-time factor, 140 MB running memories usage, 0.03 MB hard disk requirements (Table \ref{simulator_compare})
    \item[$\circ$] Reinforcement Learning Application Programming Interface (RL-API)\footnote{https://gymnasium.farama.org/api/env/} with conversion-free data flow: implements \textit{step()} and \textit{reset()} with \textit{torch.tensor}\footnotemark[2] data flow
    \item[$\circ$] Cross-platform compatibility: supports GPU/CPU execution and Ubuntu/Windows/MacOS operating systems
    \item[$\circ$] Easy usage: pure \textit{Python}\footnote{https://www.python.org} files, just \textit{import}, no installation
  \end{itemize}
\end{itemize}

\subsubsection{Conversion-free data flow}
Due to disparate implementation, most simulators to date have different data flow compared with the DRL algorithm. \textcolor{black}{That is, different data representation formats are used by the simulator and agent.} Consequently, it is inevitable to perform certain data format conversions during training. A common case is from the \textit{List} format in \textit{Python}\footnotemark[5] (generated by the simulator, stored in RAM) to the \textit{tensor} format in \textit{torch}\footnotemark[2] (utilized by the deep neural networks, stored in GPU). Such conversion is tedious and time-consuming, which slows down the training and can be even aggravated in a large-scale learning framework that requires high data throughput. To tackle this issue, we implemented the Sparrow with the prevalent machine learning framework \textit{torch}\footnotemark[2] such that the data generated by Sparrow is already stored in GPU under \textit{torch.tensor}\footnotemark[2] format, thus obviating the conversion procedure. Concurrently, to facilitate seamless integration with DRL algorithms, we provide standard RL-API, including \textit{reset()} and \textit{step()} functions, with inherent normalization integrated.

\subsubsection{Simplified kinematics}
As illustrated in Fig. \ref{s_r} (a). The kinematics of the robot in Sparrow can be described as:
\begin{equation}
	\label{knmic}
	[v^{t+1}_{l},\ v^{t+1}_{a}] = K\cdot[v^{t}_{l},\ v^{t}_{a}]+(1-K)\cdot[\hat{v}^{t}_{l},\ \hat{v}^{t}_{a}]
\end{equation}
\begin{equation}
	\label{knmic_xyphi}
	[x^{t+1}, y^{t+1}, \phi^{t+1}] = [x^{t}, y^{t}, \phi^{t}] + \Delta t \cdot [-v^{t+1}_{l} \cos(\phi^{t}),\ -v^{t+1}_{l} \sin(\phi^{t}),\ v^{t+1}_{a}]
\end{equation}

\noindent where $v$ and $\hat{v}$ respectively denote the current and target velocity of the robot, $x,y,\phi$ stand for the 2D coordinates and orientation of the robot, $\Delta t$ indicates the temporal simulation interval, the subscripts $l$ and $a$ are the respective abbreviations of linear and angular velocities, the superscripts $t$ represents the timestep, and $K$ is a hyperparameter between (0,1) that describes the combined effect of inertia, friction, and the underlying motion control algorithm. We simplify the kinematic model in this way for two reasons. First, such representation evades complicated physics calculation, conducing to the lightness of Sparrow. Second, we argue that it is inefficient or even unwise to separately investigate the coupling effects resulting from inertia, friction, and the underlying motion control algorithm due to their intricate interplay in the real world. Since these factors jointly affect how the velocity of the robot changes given the target velocity, we employed the parameter $K$ to represent the final effects. \textcolor{black}{Although $K$ cannot be measured accurately in real-world or even varies in different scenarios, we can guarantee the effectiveness of the simplified kinematics by utilizing vectorized diversity to train a generalized agent. Specifically, the parameter $K$ is randomized across different instances of the vectorized environments during training, centered around a value roughly measured from the real-world robot, which aims to foster the development of a trained agent with strong generalization capabilities over real-world operating conditions. The measurement can be achieved by tuning a $K$ value that approximately aligns with the velocity curves of the real-world robot during acceleration and deceleration, given a target velocity. Since $K$ will be randomized, an exact equivalent value corresponding to the real-world robot is not required.}
\color{black}
\subsubsection{Vectorization with world matrix}
Thanks to the simplified kinematics, an instantiation of Sparrow can be denoted as a world variable, termed $[x, y, \phi, v_l, v_a, \Delta t, K, O_{xy}]$. Here, $O_{xy}$ is the 2D coordinates of obstacle-occupied grids and will be used for LiDAR scanning. In doing so, Sparrow’s vectorized environments can be represented as a world matrix as illustrated in Fig. \ref{S&R} (a), and its parallel simulation can be realized by matrix operation, which can be efficiently addressed by the GPU. Consequently, the motion simulation of N vectorized environments can be processed via

\begin{equation}
\label{vec_vlva}
\left[\begin{array}{cc}
v_{l 1}^{t+1} & v_{a 1}^{t+1} \\
v_{l 2}^{t+1} & v_{a 2}^{t+1} \\
\vdots & \vdots \\
v_{l N}^{t+1} & v_{a N}^{t+1}
\end{array}\right]=\left[\begin{array}{cc}
K_1 & K_1 \\
K_2 & K_2 \\
\vdots & \vdots \\
K_N & K_N
\end{array}\right] \cdot\left[\begin{array}{cc}
v_{l 1}^{t} & v_{a 1}^{t} \\
v_{l 2}^{t} & v_{a 2}^{t} \\
\vdots & \vdots \\
v_{l N}^{t} & v_{a N}^{t}
\end{array}\right]+\left(\overrightarrow{1}-\left[\begin{array}{cc}
K_1 & K_1 \\
K_2 & K_2 \\
\vdots & \vdots \\
K_N & K_N
\end{array}\right]\right)  \cdot \left[\begin{array}{cc}
\hat{v}_{l 1}^{t} & \hat{v}_{a 1}^{t} \\
\hat{v}_{l 2}^{t} & \hat{v}_{a 2}^{t} \\
\vdots & \vdots \\
\hat{v}_{l N}^{t} & \hat{v}_{a N}^{t}
\end{array}\right]
\end{equation}
\begin{equation}
\label{vec_xyphi}
\left[\begin{array}{ccc}
x_1^{t+1} & y_1^{t+1} & \phi_1^{t+1} \\
x_2^{t+1} & y_2^{t+1} & \phi_2^{t+1} \\
\vdots & \vdots & \vdots \\
x_N^{t+1} & y_N^{t+1} & \phi_N^{t+1}
\end{array}\right]=\left[\begin{array}{ccc}
x_1^{t} & y_1^{t} & \phi_1^{t} \\
x_2^{t} & y_2^{t} & \phi_2^{t} \\
\vdots & \vdots & \vdots \\
x_N^{t} & y_N^{t} & \phi_N^{t}
\end{array}\right]+\Delta t\left[\begin{array}{ccc}
v_{l 1}^{t+1} & v_{l 1}^{t+1} & v_{a 1}^{t+1} \\
v_{l 2}^{t+1} & v_{l 2}^{t+1} & v_{a 2}^{t+1} \\
\vdots & \vdots & \vdots \\
v_{l N}^{t+1} & v_{l N}^{t+1} & v_{a N}^{t+1}
\end{array}\right] \cdot\left[\begin{array}{ccc}
-\cos \left(\phi_1^{t}\right) & -\sin \left(\phi_1^{t}\right) & 1 \\
-\cos \left(\phi_2^{t}\right) & -\sin \left(\phi_2^{t}\right) & 1 \\
\vdots & \vdots & \vdots \\
-\cos \left(\phi_N^{t}\right) & -\sin \left(\phi_N^{t}\right) & 1
\end{array}\right]
\end{equation}
\noindent where $\cdot$ is element-wise product.

Regarding the parallel LiDAR scanning across vectorized environments, the variability in obstacle-occupied grids across different maps results in varying lengths of $O_{xy}$. To facilitate matrix operations, $O_{xy1}, O_{xy2}, ..., O_{xyN}$ are padded to their maximum length before concatenation into a matrix. Meanwhile, the ray casting algorithm has been extended to a vectorized fashion to support parallel LiDAR scanning within vectorized environments. As this part goes beyond the core contribution of this paper, and a comprehensive elucidation can take up significant space, interested readers are recommended to consult our code for detailed implementation. 

\begin{figure}[t]
	\centering
	\subfloat[]{\includegraphics[width=0.333\textwidth]{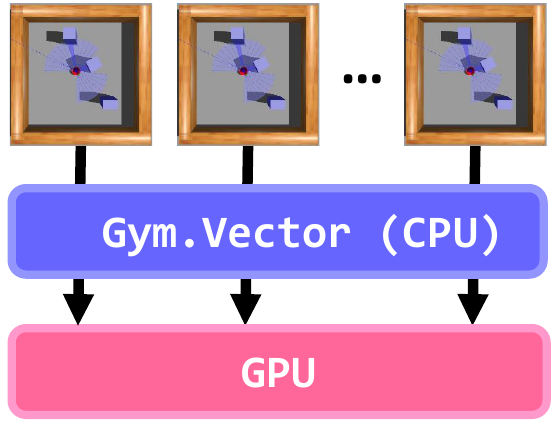}
		\label{parallel_gym}}
	\hfil
	\subfloat[]{\includegraphics[width=0.223\textwidth]{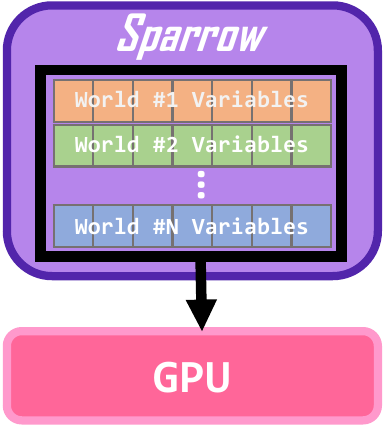}
		\label{parallel_sparrow}}
	\caption{Vectorization comparison between (a) gym and (b) Sparrow}
	\label{parallel_compare}
\end{figure}

Currently, prevalent robot simulators, such as Gazebo and Webots, lack inherent vectorization support. A common and feasible approach for their parallel simulation involves utilizing \textit{gym.vector}\footnote{https://gymnasium.farama.org/api/vector/} developed by OpenAI, as illustrated in \mbox{Fig. \ref{parallel_compare} (a)}. However, this approach exhibits notable limitations: (i) \textit{gym.vector} fundamentally functions as a multi-process management tool, wherein each process's simulation requires asynchronous calls to the GPU, resulting in pseudo-parallelism. (ii) The necessity of implementing additional mechanisms for managing shared data across many processes substantially increases communication overhead. (iii) The absence of support for shared management of GPU data necessitates data conversion between the CPU and GPU, incurring additional computational costs.
In contrast, the vectorization with Sparrow's world matrix, as illustrated in Fig. \ref{parallel_compare} (b), achieves parallel simulation of multiple environment copies through matrix operations. This methodology renders the time complexity of parallel simulation independent of the number of environments N under the GPU setting, realizing genuine parallel simulation. Meanwhile, the matrix-represented environment data obviates the need for data-sharing management and can be executed directly on GPUs, significantly reducing computational overhead. Furthermore, owing to Sparrow's lightweight design, hardware resources are not excessively consumed even with extensive vectorization. This significantly expedites data collection by aggregating data from more environmental copies under equivalent computational resources.

\subsubsection{“Colorful” training data with vectorized diversity}
Another noteworthy benefit brought about by vectorization is vectorized diversity. That is, diversified simulation setups could be adopted by different copies of the vectorized environments and be simulated simultaneously. Specifically, the world variable $[x, y, \phi, v_l, v_a, \Delta t, K, O_{xy}]$, along with the magnitude of sensor noise, can be diversified across N distinct copies of vectorized environments. Consequently, each copy represents a unique working condition, simulating various real-world scenarios such as varying illumination, battery levels, friction coefficients, mass distributions, and map layouts. This approach facilitates the generation of “colorful” training learning data, thus potentially enhancing the DRL agent's generalization capability \citep{DR}. An intuitive example is that the real-world working condition under test can be regarded as one of the N diversified simulation setups. It is important to emphasize that the diversified simulation setup for each environmental copy is re-randomized at the onset of each episode to maximize coverage of working conditions, rather than being constrained by N.
\color{black}

\subsubsection{AutoReset}
\begin{figure}[H]
	\centering
	\includegraphics[width=0.99\textwidth]{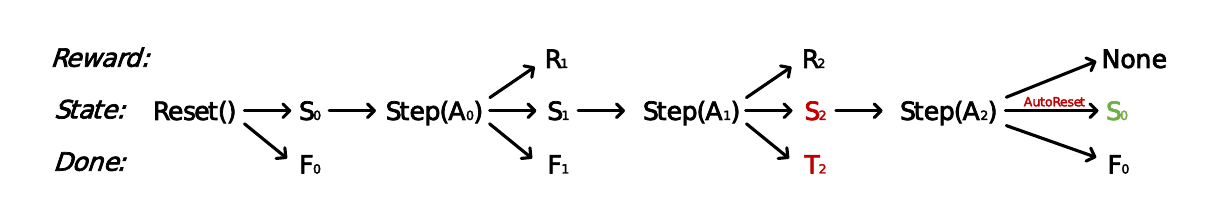}
	\caption{AutoReset reset mechanism}
	\label{AutoReset}
\end{figure}

The environment copies within the vectorized environment may be \textit{done} (terminated or truncated) at disparate timesteps. Consequently, invoking the \textit{reset()} function for all copies upon the \textit{done} signal of a single copy proves inefficient and even inappropriate, necessitating the development of an advanced reset mechanism. To address this, Sparrow incorporates an AutoReset mechanism that implicitly resets the \textit{done} copy within the \textit{step()} function without affecting other copies, as illustrated in Fig. \ref{AutoReset}.

\color{black}

\subsection{Color}
Having comprehended the ASL framework and the Sparrow simulator, the only remaining question pertains to their optimal amalgamation. That is, forming a comprehensive solution that enables the efficient training of a DRL agent with adequate resilience to generalize over a series of analogous scenarios as well as real-world scenarios. Our method, named Color, is illustrated in Fig. \ref{Color}, wherein the ASL and the Sparrow are seamlessly integrated by their interdependent vectorized environments, and the simulation setups are randomized in different copies of the vectorized environments of Sparrow. Through interacting with the vectorized and diverse environments, the actor of ASL is empowered to gather a collection of ``colorful'' training data, which is then preserved in the sharer and prepared for the model optimization of the learner. In this fashion, the training efficiency of Color is guaranteed by the high data throughput of Sparrow and the efficiency of ASL, while the generalization capacity is promised by learning from the ``colorful'' training data. 

\section{Experiments}
\label{section4}
\begin{figure}[H]
	\centering
	\includegraphics[width=0.5\textwidth]{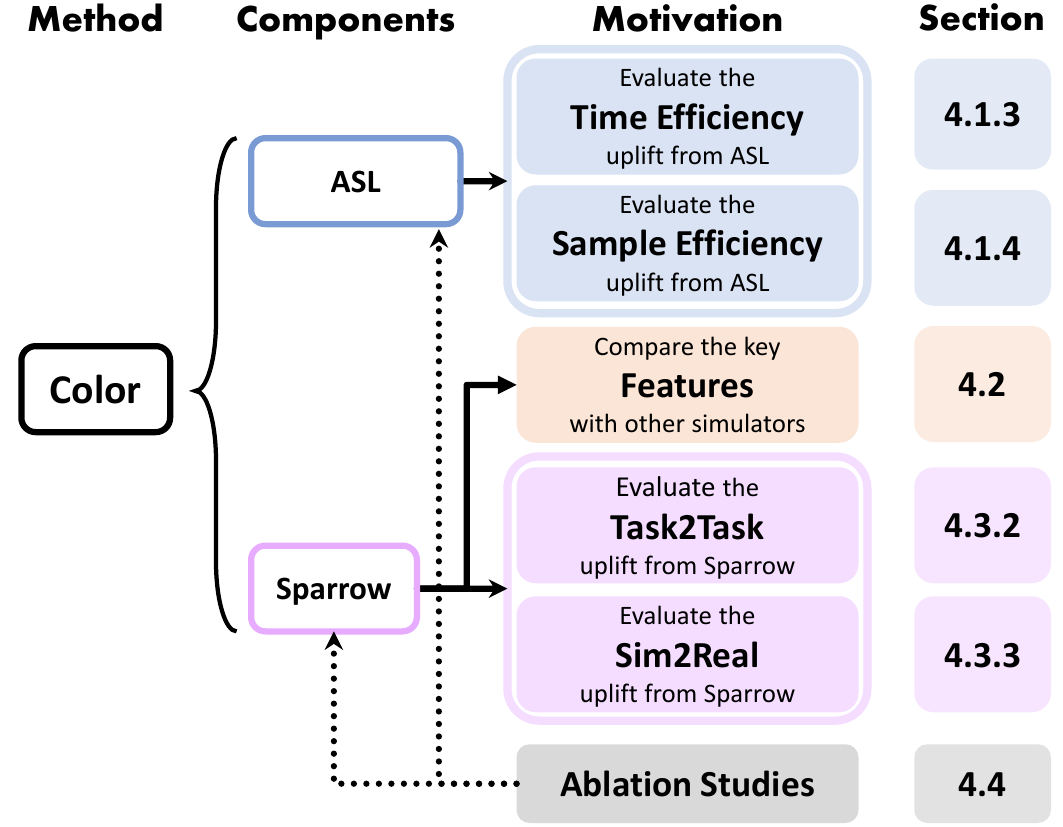}
	\caption{Arrangement of experiments.}
	\label{exp_arangement}
\end{figure}

As illustrated in Fig. \ref{exp_arangement}, this section conducts four distinct experiments to substantiate the superiority of the proposed methods. The first experiment focuses on the efficiency of the ASL framework. Comprehensive experiments have been conducted on 57 Atari games \citep{atari}, and the results are compared with other previously published DRL baselines. The second experiment involves a comparative study of Sparrow and other prevalent simulators. The third experiment examines the Task2Task and Sim2Real capabilities of the agent derived from Color. The final experiment ablates the key components of Color to determine their respective contributions. \textcolor{black}{The hardware and software configurations supporting these experiments are detailed in Table \ref{tab:hp} $\sim$ \ref{tab:pythonlib} for reproducibility.}

\subsection{Evaluation of ASL on Atari environments}
\textcolor{black}{In this section, we evaluate the proposed ASL training framework on benchmark Atari environments, assessing its training time efficiency, sample efficiency, stability, and final performance.}

\color{black}
\subsubsection{Atari environments}

\begin{figure}[H]
	\centering
	\includegraphics[width=0.8\textwidth]{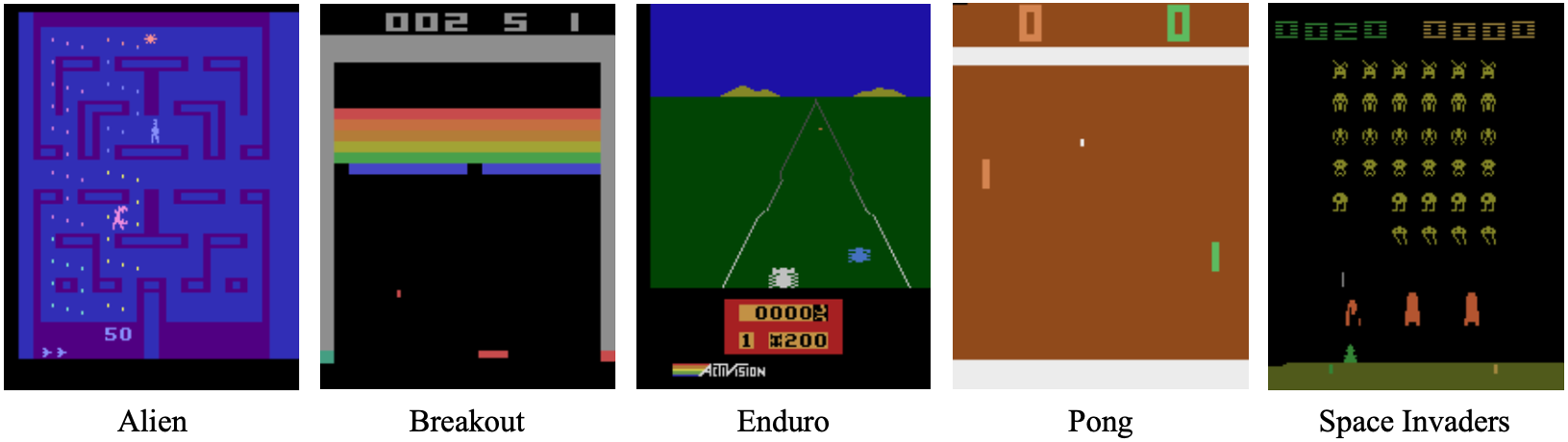}
	\caption{\textcolor{black}{Representative games of the Atari 57 environment.}}
	\label{atari_imgs}
\end{figure}

The Atari 57 environment \citep{atari}, illustrated in Fig. \ref{atari_imgs}, consists of a comprehensive suite of benchmark tasks based on classic Atari 2600 video games. These tasks present diverse challenges, evaluating an agent's ability to learn from high-dimensional raw pixel inputs and make decisions accordingly. The tasks vary in complexity, combining control, exploration, and strategic elements, and serve as a rigorous platform for researchers to evaluate and compare DRL algorithms and frameworks.

\color{black}

\subsubsection{Experiment setups and results on Atari}
To validate the efficiency of ASL, its results on 57 Atari games are compared with DQN \citep{DQN}, DDQN \citep{DDQN}, Prior. DDQN \citep{PER}, Duel. DDQN \citep{duel}, Noisy DQN \citep{noisy} and DSQN \citep{DSQN}. Essentially, ASL is a DRL training framework and should be applied in conjunction with a specific DRL algorithm. Although extensive improvement techniques have emerged since the ground-breaking work of DQN, we choose to combine the ASL framework merely with the DDQN, an algorithm that mitigates the notorious overestimation problem of DQN by Double Q-learning, and omit other tricks such as PER, dueling network, noisy network, etc. We do this with the intention of preventing these techniques from submerging the ASL. In this context, we refer to our algorithm as ASL DDQN. In addition, since the ASL is mostly inspired by the Ape-X framework, we also compare our algorithm with the Ape-X DDQN. For a fair comparison, we follow the experimental setup of previous works. We employ the same Atari preprocess procedures and model architecture of DDQN. Furthermore, we also adopt the no-ops scheme, which imposes a random number (upper-bounded by 30) of no-op actions at the beginning of each episode. The no-ops scheme could effectively prevent the agent from overfitting when training while also examine its robustness when evaluating. In addition, the episode of each Atari game is terminated at 12.5K steps and 27K steps for training and evaluation, respectively. 

Regarding the hyperparameters, since the ASL DDQN is constituted of a training framework and a DRL algorithm, the hyperparameters consist of two parts as well. Considering the prohibitively high cost of fulfilling an exhaustive search on the combined hyperparameter space, we determine the algorithm-associated hyperparameters based on the published papers and the framework-relevant hyperparameters mostly by manual coordinate descent. For instance, we did not observe a noticeable impact of the model upload frequency $U$ and therefore made an accommodation between the model recency and transmission cost. In addition, the number of vectorized environments $N$ is assigned a value of 128 to correspond with the maximum capacity of our CPU (AMD 3990X, 64 cores, 128 threads). The $\mbox{\textit{TPS}}$ is set to 8, the same as DDQN (a backpropagation with a mini-batch of size 32 every after 4 transitions have been collected). More information about the hyperparameter is listed in Table \ref{tab:hp}. Note that the hyperparameters across all 57 Atari games are identical.

\textcolor{black}{The ASL DDQN is trained for 50M $T_{step}$, adhering to the experimental setups within the DRL community \citep{DQN,DDQN,PER,duel,noisy,DSQN}. Meanwhile, the Ape-X DDQN is trained for 500M $T_{step}$ so that the overall training time of Ape-X is roughly the same as ASL.} During the training, the model is evaluated every 5K $B_{step}$, with Fig. \ref{t_e} and Fig. \ref{s_e} from the Appendix comparing the ASL DDQN and Ape-X DDQN in terms of sample efficiency and training time efficiency, respectively. \color{black}Furthermore, we also introduce the Sample Efficiency Comparison (SEC) and Training Time Efficiency Comparison (TTEC) metrics to quantify the improvements of ASL over Ape-X:
\newpage

\begin{equation}
	\label{SEC}
	\text{SEC} = \frac{1}{n}\sum_{i=1}^{n}\frac{\text{Sample}^{\text{ASL}}(S_i)}{\text{Sample}^{\text{Ape-X}}(S_i)}
\end{equation}

\begin{equation}
	\label{TTEC}
	\text{TTEC} = \frac{1}{n}\sum_{i=1}^{n}\frac{\text{Time}^{\text{ASL}}(S_i)}{\text{Time}^{\text{Ape-X}}(S_i)}
\end{equation}

\noindent where 
$i$ is the index of the Atari games, 
$n$ (57) is the number of Atari games.
For the i-th Atari game, 
let $S_i$ denote the minimum of the final scores achievable by Ape-X and ASL.
Here, we employ the minimum one to ensure the accessibility of both algorithms. 
$\text{Sample}^{\text{Ape-X}}(S_i)$ and $\text{Sample}^{\text{ASL}}(S_i)$ represent the number of training samples required by Ape-X and ASL, respectively, to attain $S_i$. 
Similarly, $\text{Time}^{\text{Ape-X}}(S_i)$ and $\text{Time}^{\text{ASL}}(S_i)$ denote the training time necessary for Ape-X and ASL, respectively, to reach Si.
As a result, SEC and TTEC compare the relative training sample and time requirements of ASL and Ape-X to achieve equivalent performance. To mitigate the impact of outliers and better represent overall trends, five minimum and maximum values were excluded from the 57 Atari games. The resulting data are then summarized in Table \ref{tab:SETE}.
\color{black}

\begin{table}[H]
\center
\caption{Sample and training time efficiency comparison between ASL and Ape-X}
\label{tab:SETE}%
\begin{tabular}{lll}
		\hline
		\textbf{Compared Algorithm} & \textbf{SEC} & \textbf{TTEC} \\
		\hline
		ASL v.s. Ape-X & 6.02\% & 46.47\% \\
		\hline
	\end{tabular}%
\end{table}%

\color{black}
After training, the best models of the 57 Atari games are evaluated for 100 episodes, and the averaged raw scores are compared with other DRL baselines published previously, as the results presented in Table \ref{tab:a57}.

\begin{table}
	\centering
	\caption{\textcolor{black}{Raw Score Comparison Across 57 Atari Games}}
        \begin{adjustbox}{center}
	\resizebox{1.25\textwidth}{!}{
		\begin{tabular}{lrrrrrrrrrrr}
			\hline
			\textbf{Game} & \textbf{Random} & \textbf{DQN} & \textbf{DDQN} & \textbf{Prior. DDQN} & \textbf{Duel. DDQN} & \textbf{Noisy DQN} & \textbf{DSQN} & \textbf{Ape-X DDQN} & \textbf{ASL DDQN (Ours)} & \textbf{Ipv1} & \textbf{Ipv2} \\
			\hline
			Alien & 227.8 & 3,069.0 & 2,907.3 & 4,203.8 & 4,461.4 & 2,403.0 & -     & 2,596.1 & \textbf{     6,955.2 } & 251.1\% & 284.1\% \\
			Amidar & 5.8   & 739.5 & 702.1 & 1,838.9 & \textbf{2,354.5} & 1,610.0 & -     & 1,271.2 &      2,232.3  & 319.8\% & 176.0\% \\
			Assault & 222.4 & 3,359.0 & 5,022.9 & 7,672.1 & 4,621.0 & 5,510.0 & -     & \textbf{14,490.1} &     14,372.8  & 294.8\% & 99.2\% \\
			Asterix & 210.0 & 6,012.0 & 15,150.0 & 31,527.0 & 28,188.0 & 14,328.0 & -     & 189,300.0 & \textbf{ 567,640.0 } & 3798.1\% & 300.1\% \\
			Asteroids & 719.1 & 1,629.0 & 930.6 & 2,654.3 & 2,837.7 & \textbf{3,455.0} & -     & 1,107.9 &      1,984.5  & 598.3\% & 325.5\% \\
			Atlantis & 12,850.0 & 85,641.0 & 64,758.0 & 357,324.0 & 382,572.0 & 923,733.0 & 487,366.7 & 831,952.0 & \textbf{ 947,275.0 } & 1800.2\% & 114.1\% \\
			BankHeist & 14.2  & 429.7 & 728.3 & 1,054.6 & \textbf{1,611.9} & 1,068.0 & -     & 1,264.9 &      1,340.9  & 185.8\% & 106.1\% \\
			BattleZone & 2,360.0 & 26,300.0 & 25,730.0 & 31,530.0 & 37,150.0 & 36,786.0 & -     & 38,671.0 & \textbf{   38,986.0 } & 156.7\% & 100.9\% \\
			BeamRider & 363.9 & 6,846.0 & 7,654.0 & 23,384.2 & 12,164.0 & 20,793.0 & 7,226.9 & \textbf{27,033.7} &     26,841.6  & 363.2\% & 99.3\% \\
			Berzerk & 123.7 & -     & -     & 1,305.6 & 1,472.6 & 905.0 & -     & 924.8 & \textbf{     2,597.2 } & -     & 308.8\% \\
			Bowling & 23.1  & 42.4  & 70.5  & 47.9  & 65.5  & \textbf{71.0} & -     & 60.2  &           62.4  & 82.9\% & 105.9\% \\
			Boxing & 0.1   & 71.8  & 81.7  & 95.6  & 99.4  & 89.0  & 95.3  & 99.4  & \textbf{         99.6 } & 121.9\% & 100.2\% \\
			Breakout & 1.7   & 401.2 & 375.0 & 373.9 & 345.3 & 516.0 & 386.5 & 370.7 & \textbf{       621.7 } & 166.1\% & 168.0\% \\
			Centipede & 2,090.9 & \textbf{8,309.0} & 4,139.4 & 4,463.2 & 7,561.4 & 4,269.0 & -     & 3,808.4 &      3,899.8  & 88.3\% & 105.3\% \\
			ChopperCommand & 811.0 & 6,687.0 & 4,653.0 & 8,600.0 & 11,215.0 & 8,893.0 & -     & 6,031.0 & \textbf{   15,071.0 } & 371.2\% & 273.2\% \\
			CrazyClimber & 10,780.5 & 14,103.0 & 101,874.0 & 141,161.0 & 143,570.0 & 118,305.0 & 123,916.7 & 118,020.0 & \textbf{ 166,019.0 } & 170.4\% & 144.8\% \\
			Defender & 2,874.5 & -     & -     & 31,286.5 & \textbf{42,214.0} & 20,525.0 & -     & 29,255.0 &     37,026.5  & -     & 129.5\% \\
			DemonAttack & 152.1 & 9,711.0 & 9,711.9 & 71,846.4 & 60,813.3 & 36,150.0 & -     & 114,874.7 & \textbf{ 119,773.9 } & 1251.3\% & 104.3\% \\
			DoubleDunk & -18.6 & -18.1 & -6.3  & \textbf{18.5} & 0.1   & 1.0   & -     & -0.2  &             0.1  & 152.0\% & 101.6\% \\
			Enduro & 0.0   & 301.8 & 319.5 & 2,093.0 & \textbf{2,258.2} & 1,240.0 & -     & 1,969.1 &      2,103.1  & 658.2\% & 106.8\% \\
			FishingDerby & -91.7 & -0.8  & 20.3  & 39.5  & \textbf{46.4} & 11.0  & -     & 31.2  &           35.1  & 113.2\% & 103.2\% \\
			Freeway & 0.0   & 30.3  & 31.8  & 33.7  & 0.0   & 32.0  & -     & 21.4  & \textbf{         33.9 } & 106.6\% & 158.4\% \\
			Frostbite & 65.2  & 328.3 & 241.5 & 4,380.1 & 4,672.8 & 753.0 & -     & 504.7 & \textbf{     8,616.4 } & 4850.4\% & 1945.7\% \\
			Gopher & 257.6 & 8,520.0 & 8,215.4 & 32,487.2 & 15,718.4 & 14,574.0 & 10,107.3 & 47,845.6 & \textbf{ 103,514.4 } & 1297.6\% & 217.0\% \\
			Gravitar & 173.0 & 306.7 & -     & 548.5 & 588.0 & 447.0 & -     & 242.5 & \textbf{       760.0 } & -     & 844.6\% \\
			Hero  & 1,027.0 & 19,950.0 & 20,357.0 & 23,037.7 & 20,818.2 & 6,246.0 & -     & 14,464.0 & \textbf{   26,578.5 } & 132.2\% & 190.2\% \\
			IceHockey & -11.2 & -1.6  & -2.4  & \textbf{1.3} & 0.5   & -3.0  & -     & -2.5  &           -3.6  & 86.4\% & 87.4\% \\
			Jamesbond & 29.0  & 576.7 & 438.0 & \textbf{5,148.0} & 1,312.5 & 1,235.0 & 1,156.7 & 540.0 &      2,237.0  & 539.9\% & 432.1\% \\
			Kangaroo & 52.0  & 6,740.0 & 13,651.0 & \textbf{16,200.0} & 14,854.0 & 10,944.0 & 8,880.0 & 14,710.0 &     13,027.0  & 95.4\% & 88.5\% \\
			Krull & 1,598.0 & 3,805.0 & 4,396.7 & 9,728.0 & \textbf{11,451.9} & 8,805.0 & 9,940.0 & 10,999.4 &     10,422.5  & 315.3\% & 93.9\% \\
			KungFuMaster & 258.5 & 23,270.0 & 29,486.0 & 39,581.0 & 34,294.0 & 36,310.0 & -     & 54,124.0 & \textbf{   85,182.0 } & 290.6\% & 157.7\% \\
			MontezumaRevenge & 0.0   & 0.0   & 0.0   & 0.0   & 0.0   & \textbf{3.0} & -     & 0.0   & 0.0   & -     & - \\
			MsPacman & 307.3 & 2,311.0 & 3,210.0 & \textbf{6,518.7} & 6,283.5 & 2,722.0 & -     & 4,087.7 &      4,416.0  & 141.5\% & 108.7\% \\
			NameThisGame & 2,292.3 & 7,257.0 & 6,997.1 & 12,270.5 & 11,971.1 & 8,181.0 & 10,877.0 & 16,042.7 & \textbf{   16,535.4 } & 302.7\% & 103.6\% \\
			Phoenix & 761.4 & -     & -     & 18,992.7 & 23,092.2 & 16,028.0 & -     & 28,296.0 & \textbf{   71,752.6 } & -     & 257.8\% \\
			Pitfall & -229.4 & -     & -     & -356.5 & \textbf{0.0}   & \textbf{0.0}   & -     & \textbf{0.0}   & \textbf{0.0}   & -     & 100.0\% \\
			Pong  & -20.7 & 18.9  & \textbf{21.0} & 20.6  & \textbf{21.0} & \textbf{21.0} & 20.3  & \textbf{21.0} & \textbf{         21.0 } & 100.0\% & 100.0\% \\
			PrivateEye & 24.9  & 1,788.0 & 670.1 & 200.0 & 103.0 & \textbf{3,712.0} & -     & 173.0 &         349.7  & 50.3\% & 219.3\% \\
			Qbert & 163.9 & 10,596.0 & 14,875.0 & 16,256.5 & 19,220.3 & 15,545.0 & -     & 15,300.0 & \textbf{   24,548.8 } & 165.8\% & 161.1\% \\
			Riverraid & 1,338.5 & 8,316.0 & 12,015.3 & 14,522.3 & 21,162.6 & 9,425.0 & -     & 22,238.0 & \textbf{   24,445.0 } & 216.4\% & 110.6\% \\
			RoadRunner & 11.5  & 18,257.0 & 48,377.0 & 57,608.0 & \textbf{69,524.0} & 45,993.0 & 48,983.3 & 51,208.0 &     56,520.0  & 116.8\% & 110.4\% \\
			Robotank & 2.2   & 51.6  & 46.7  & 62.6  & 65.3  & 51.0  & -     & 42.6  & \textbf{         65.8 } & 142.9\% & 157.4\% \\
			Seaquest & 68.4  & 5,286.0 & 7,995.0 & 26,357.8 & \textbf{50,254.2} & 2,282.0 & -     & 32,101.8 &     29,278.6  & 368.5\% & 91.2\% \\
			Skiing & -17,098.1 & -     & -     & -9,996.9 & -8,857.4 & -14,763.0 & -     & -15,623.8 &     \textbf{-8,295.4}  & -     & 597.1\% \\
			Solaris & 1,236.3 & -     & -     & 4,309.0 & 2,250.8 & \textbf{6,088.0} & -     & 1,523.2 &      3,506.8  & -     & 791.4\% \\
			SpaceInvaders & 148.0 & 1,976.0 & 3,154.6 & 2,865.8 & 6,427.3 & 2,186.0 & 1,832.2 & 3,943.0 & \textbf{   21,602.0 } & 713.6\% & 565.3\% \\
			StarGunner & 664.0 & 57,997.0 & 65,188.0 & 63,302.0 & 89,238.0 & 47,133.0 & 57,686.7 & 60,835.0 & \textbf{ 129,140.0 } & 199.1\% & 213.5\% \\
			Surround & -10.0 & -     & -     & \textbf{8.9} & 4.4   & -1.0  & -     & 2.9   &             2.5  & -     & 96.9\% \\
			Tennis & -23.8 & -2.5  & 1.7   & 0.0   & 5.1   & 0.0   & -1.0  & -0.9  & \textbf{         22.3 } & 180.8\% & 201.3\% \\
			TimePilot & 3,568.0 & 5,947.0 & 7,964.0 & 9,197.0 & 11,666.0 & 7,035.0 & -     & 7,457.0 & \textbf{   12,071.0 } & 193.4\% & 218.6\% \\
			Tutankham & 11.4  & 186.7 & 190.6 & 204.6 & 211.4 & 232.0 & 194.7 & 226.2 & \textbf{       252.9 } & 134.8\% & 112.4\% \\
			UpNDown & 533.4 & 8,456.0 & 16,769.9 & 16,154.1 & 44,939.6 & 14,255.0 & -     & \textbf{46,208.0} &     25,127.4  & 151.5\% & 53.8\% \\
			Venture & 0.0   & 380.0 & 93.0  & 54.0  & \textbf{497.0} & 97.0  & -     & 62.0  &         291.0  & 312.9\% & 469.4\% \\
			VideoPinball & 16,256.9 & 42,684.0 & 70,009.0 & 282,007.3 & 98,209.5 & 322,507.0 & 275,342.8 & 603,075.0 & \textbf{ 626,794.0 } & 1135.8\% & 104.0\% \\
			WizardOfWor & 563.5 & 3,393.0 & 5,204.0 & 4,802.0 & 7,855.0 & 9,198.0 & -     & 14,780.0 & \textbf{   21,049.0 } & 441.5\% & 144.1\% \\
			YarsRevenge & 3,092.9 & -     & -     & 11,357.0 & \textbf{49,622.1} & 23,915.0 & -     & 13,178.5 &     29,231.9  & -     & 259.2\% \\
			Zaxxon & 32.5  & 4,977.0 & 10,182.0 & 10,469.0 & 12,944.0 & 6,920.0 & -     & -     & \textbf{   16,420.0 } & 161.5\% & - \\
			\hline
			\multicolumn{12}{l}{Scores of Random, Prior. DDQN, and Duel. DDQN are taken from \citet{duel}.  }\\
			\multicolumn{12}{l}{Scores of DQN, Noisy DQN, DDQN, and DSQN are taken from \citet{DQN, DDQN, noisy, DSQN}, respectively. }\\
			\multicolumn{12}{l}{Since Ape-X did not publish its score combining with DDQN, the scores of Ape-X DDQN and ASL DDQN are both obtained by our experiments. }\\
			\multicolumn{12}{l}{\textbf{Ipv1} = (ASL DDQN - Random)/(DDQN - Random): the improvement of ASL DDQN over its underlying DRL algorithm DDQN.}\\
			\multicolumn{12}{l}{\textbf{Ipv2} = (ASL DDQN - Random)/(Ape-X DDQN - Random): the improvement of ASL DDQN over Ape-X DDQN.}\\
	\end{tabular} }
        \end{adjustbox}
	\label{tab:a57}%
\end{table}%

\subsubsection{Training time efficiency of ASL}
Fig. \ref{t_e} indicates that ASL surpasses Ape-X in terms of training time efficiency, achieving equivalent or superior performance with less training time across most Atari games. 
\color{black}
Specifically, Table \ref{tab:SETE} shows that ASL requires only 46.47\% of the training time compared to Ape-X to achieve comparable performance. We attribute ASL's superior training time efficiency to its VDC and partially decoupled training modes. The VDC mode maximizes GPU parallel processing capabilities and simplifies data preservation, thereby enhancing computational efficiency and reducing training duration. Additionally, the partially decoupled training mode optimizes training time by simultaneously conducting data collection and model training. Moreover, the TFM within the partially decoupled training mode ensures prudent data recency, resulting in comparable performance with fewer training instances and thus decreasing overall required training time.
\color{black}

\subsubsection{Sample efficiency of ASL}
Fig. \ref{s_e} illustrates that ASL substantially improved the sample efficiency over Ape-X. The result in Table \ref{tab:SETE} further corroborates such observation, where ASL achieves equivalent performance using only 6.02\% of the training data required by Ape-X. The improvement originates from ASL’s TFM and VEM. The relative speed of the Actor and Learner is reasonably coordinated by the TFM in accordance with the $\mbox{\textit{TPS}}$, preventing the Actor from generating excessive data. Meanwhile, the VEM overcomes the limitation of canonical $\epsilon$-greedy exploration, reaching a more rational compromise between exploration and exploitation and yielding higher quality training data that fosters training.
\color{black}

\subsubsection{Stability of ASL}
\begin{figure}[H]
    \adjustbox{width=1.26\textwidth, center}{\includegraphics{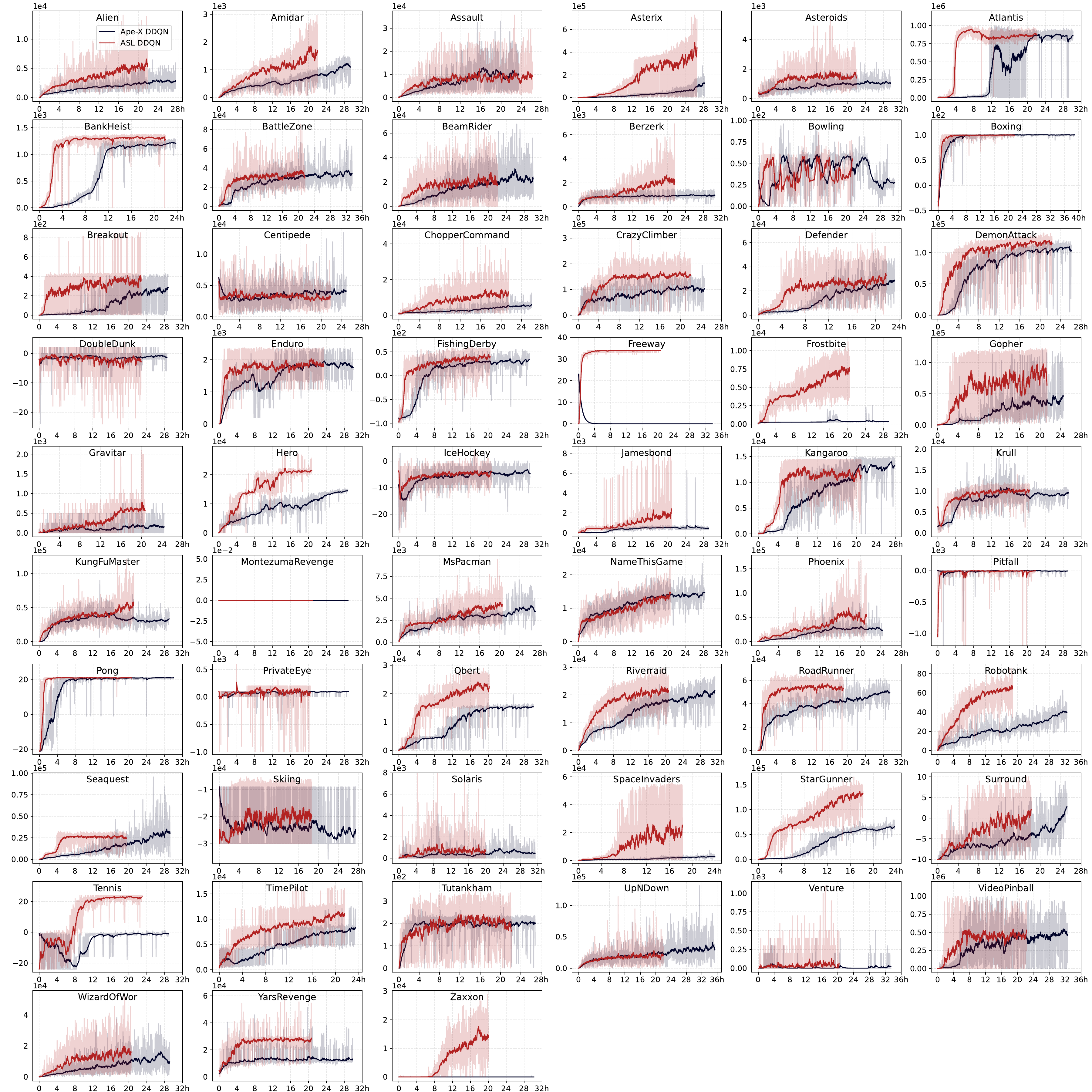}}
    \caption{Training time efficiency comparison between ASL DDQN and Ape-X DDQN. The horizontal axis is the training time, and the vertical axis corresponds to the episode reward of a single evaluation. The translucent curves represent the raw data, while the solid curves are exponentially smoothed by a factor of 0.95.}
    \label{t_e}
\end{figure}

\begin{figure}[H]
    \adjustbox{width=1.26\textwidth, center}{\includegraphics{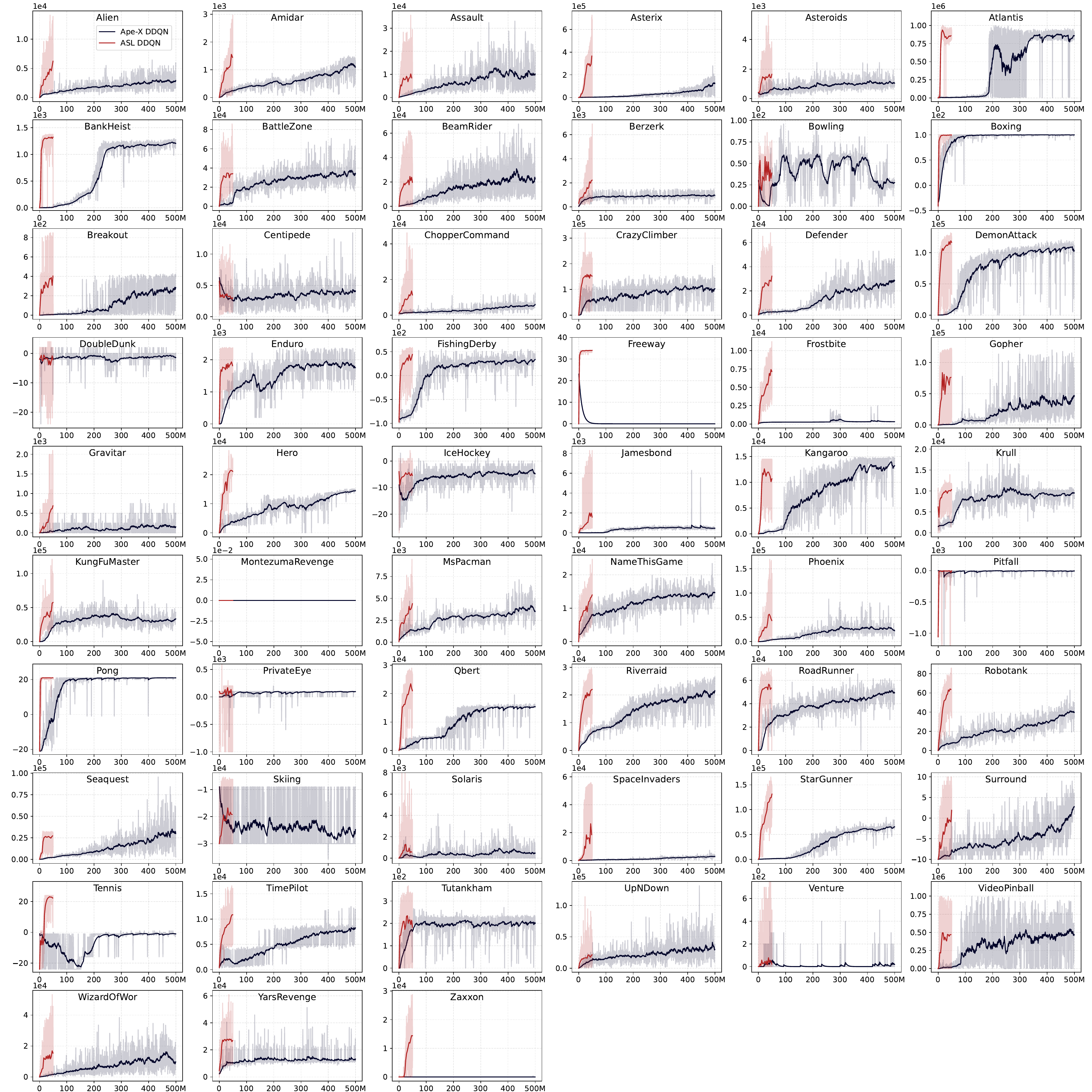}}
    \caption{Sample efficiency comparison between ASL DDQN and Ape-X DDQN. The horizontal axis is the number of transitions used during training, and the vertical axis is the episode reward of a single evaluation. The translucent curves represent the raw data, while the solid curves are exponentially smoothed by a factor of 0.95. It should be noted that this figure is derived from same experiments as Fig. \ref{t_e}, differing only in the horizontal axis representation.}
    \label{s_e}
\end{figure}

The ASL framework exhibits better stability than Ape-X, which is supported by the curves of \textit{Freeway}, \textit{Frostbite}, and \textit{Zaxxon} \textcolor{black}{from Fig. \ref{t_e} and Fig. \ref{s_e}}, where the Ape-X DDQN failed to learn. We conjecture such failures are mostly incurred by the uncontrolled $\mbox{\textit{TPS}}$ in Ape-X.

\subsubsection{Final performance of ASL}
Table \ref{tab:a57} reveals that the ASL DDQN outstrips other DRL baselines over 32 Atari games. Besides, the final performance gain achieved by the ASL DDQN over its underlying algorithm DDQN is a remarkable 508.2\% (averaged over \textit{Ipv1} across 57 Atari games). Furthermore, compared to the framework counterpart Ape-X, the ASL demonstrates a substantial improvement of 234.9\% (averaged over \textit{Ipv2} across 57 Atari games). These findings provide strong evidence of the superiority of our ASL training framework.

\subsection{Evaluation of Sparrow}
\begin{figure}[H]
	\centering
	\subfloat[]{\includegraphics[width=0.2\textwidth]{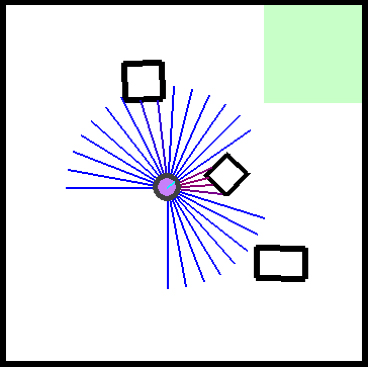}}
	\hfil
	\subfloat[]{\includegraphics[width=0.205\textwidth]{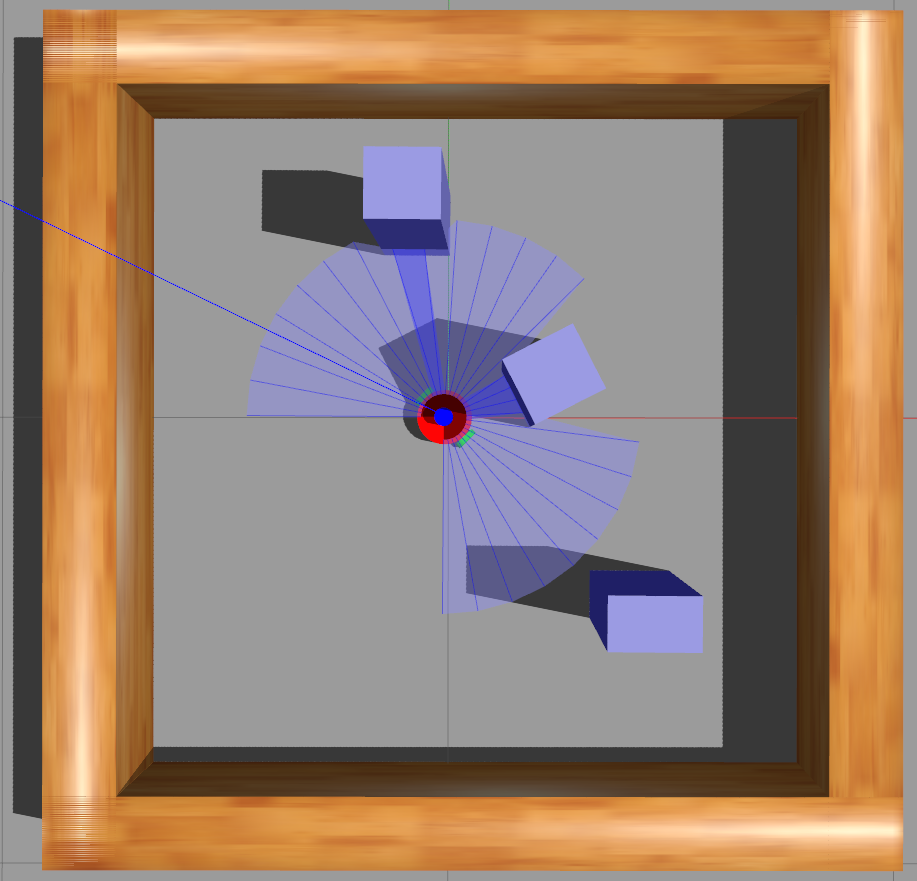}}
	\hfil
	\subfloat[]{\includegraphics[width=0.205\textwidth]{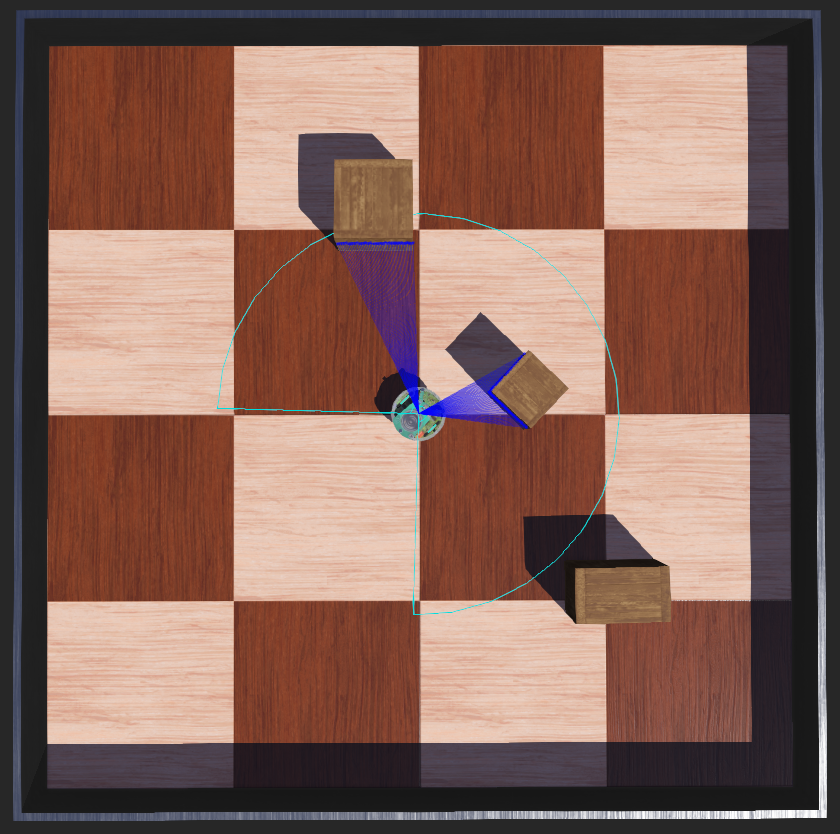}}
	\caption{Mobile robot-oriented simulators: (a) Sparrow, (b) Gazebo, (c) Webots.}
	\label{simulators}
\end{figure}

To demonstrate the superiority of Sparrow, it is compared with four wildly adopted DRL simulators. For a fair comparison, analogous layouts are constructed within the mobile robot-oriented simulators, as shown in Fig. \ref{simulators}. The comparison results are presented in Table \ref{simulator_compare}. 

\begin{table}[H]
\footnotesize
\caption{Simulator Comparison}
\label{simulator_compare}
\resizebox{\textwidth}{!}{
\begin{tabular}{ccccccccc}
\hline
\textbf{Simulator} & \textbf{Vectorization} & \textbf{DR} & \textbf{RL-API} & \textbf{CFDF} & \textbf{Robot-oriented} & \textbf{Maximal RTF}  &  \textbf{GPU Memory (MB)} & \textbf{Hard Disk (MB)}\\
\hline
    \multirow{2}[0]{*}{Sparrow (Ours)} & \multirow{2}[0]{*}{\textcolor{black}{\faCheck}} & \multirow{2}[0]{*}{\textcolor{black}{\faCheck}} & \multirow{2}[0]{*}{\textcolor{black}{\faCheck}} & \multirow{2}[0]{*}{\textcolor{black}{\faCheck}}  & \multirow{2}[0]{*}{\textcolor{black}{\faCheck}} & $N$1: 24.3 & $N$1: \textbf{140} & \multirow{2}[0]{*}{\textbf{0.03}}\\
         &       &       &       &       &       &       $N$64: \textbf{815.5} & $N$64: \textbf{140}\\
    Gazebo \citep{gazebo} & \textcolor{purple}{\faTimes}     & \textcolor{purple}{\faTimes}     & \textcolor{purple}{\faTimes}     & \textcolor{purple}{\faTimes}        & \textcolor{black}{\faCheck}     & 16.2  & 180 & 170\\
    Webots \citep{Webots} & \textcolor{purple}{\faTimes}     & \textcolor{purple}{\faTimes}     & \textcolor{purple}{\faTimes}     & \textcolor{purple}{\faTimes}        & \textcolor{black}{\faCheck}     & 113.9   & 560 & 456\\
    CoinRun \citep{CoinRun} & \textcolor{black}{\faCheck}     & \textcolor{black}{\faCheck}     & \textcolor{black}{\faCheck}     & \textcolor{purple}{\faTimes}       & \textcolor{purple}{\faTimes}     & -     & - & 38.8\\
    Procgen \citep{Procgen} & \textcolor{black}{\faCheck}     & \textcolor{black}{\faCheck}     & \textcolor{black}{\faCheck}     & \textcolor{purple}{\faTimes}       & \textcolor{purple}{\faTimes}     & -     & - & 36.6\\
\hline
\multicolumn{9}{l}{DR: Domain randomization; RL-API: Reinforcement learning application programming interfaces such as \textit{reset()} and \textit{step()};}\\
\multicolumn{9}{l}{CFDF: Conversion-free data flow; RTF: Real-time factor; $N$: Number of vectorized environments.}
\end{tabular}}%
\end{table}%

As delineated in Table \ref{simulator_compare}, Sparrow encompasses a series of advantageous features, namely RL-API, Vectorization, and DR. These attributes, often overlooked by alternative mobile robot simulators, are instrumental for DRL training, as they facilitate deployment, boost the training speed, and enhance the generalization ability of the trained agent. Concurrently, Sparrow provides a conversion-free data flow, thereby eliminating the need for data conversion between GPU and RAM, a common bottleneck in training speed. Moreover, Sparrow is the most resource-efficient among the five simulators under consideration in terms of both GPU memory and hard disk usage. To quantify the simulation speed, we employ the maximal RTF, defined as the ratio of the simulated time to the real time when running the simulator at peak speed. Although Sparrow is not the fastest in the single environment setting ($N$=1), its simulation speed could be substantially promoted through vectorization. With a vectorization of 64 copies, Sparrow achieves a maximal RTF of 815.5, remarkably surpassing the other two mobile robot-oriented simulators. In summary, we posit that Sparrow has bridged the niche between two types of simulators: a) physics engine-based simulators such as Gazebo and Webots, which lack vectorization to boost data throughput and DR to enhance the generalization capability of the trained DRL agent, and b) non-physics engine-based simulators like CoinRun and Procgen, which are limited in their contribution to real-world applications of DRL.

Finally, we would like to introduce three promising applications of Sparrow. First, as well as the most discussed, Sparrow is capable of fostering the efficiency and generalization of the trained DRL algorithm by providing vectorized diversity. Second, the broad trend in the DRL community to date is evaluating the performance of agent on the training environment. We contend such evaluation is irrational to some extent because it is equivalent to evaluate on the training set in supervised learning. Thanks to the diversifiable characteristics of Sparrow, it is feasible to quantify the generalization of the DRL agent more precisely by constructing different simulation parameters or maps during evaluation. Last, the lightness and the conversion-free data flow of Sparrow render it notably user-friendly, thereby promoting the potential for expeditious deployment, assessment, and comparison of DRL algorithms. For these reasons, we believe that Sparrow could be deemed as a holistic simulation platform that would profitably contribute to the community of DRL researchers and practitioners.

\subsection{Evaluation of Color on LLP problem}
This section endeavors to investigate the generalization capabilities of the agent trained by Color. It commences with the introduction of the problem formulation of LLP. Subsequently, it assesses the Task2Task and Sim2Real capacity of the Color agent within Sparrow and real-world environments.

\subsubsection{Problem formulation of LLP}

The LPP problem is formulated as a Markov Decision Process (MDP) to enable the employment of DRL. At timestep $t$, the agent observes the state $s_t$ from the environment, takes the action $a_t$ according to its policy $\pi$, receives the reward $r_t$, and subsequently transits to the next state $s_{t+1}$, with the objective of maximizing the expected sum of discounted rewards $\mathbb{E}_\pi\left[\sum_{t=0}^{+\infty} \gamma^t r_t\right]$, where $\gamma \in[0,1]$ is the discount factor.

\begin{figure}[t]
	\centering
	\subfloat[]{\includegraphics[width=0.35\textwidth]{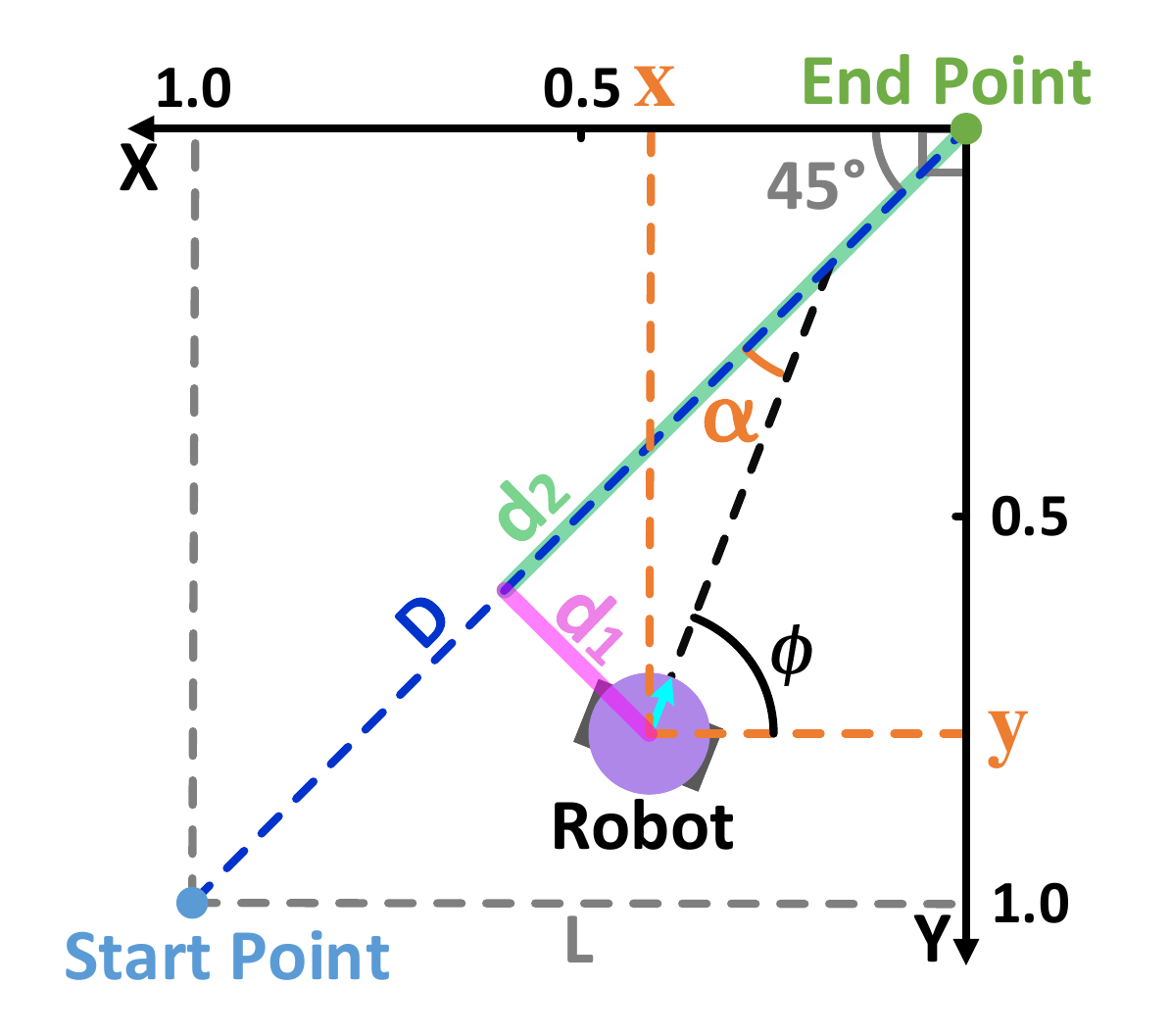}
		\label{s_train}}
	\hfil
	\subfloat[]{\includegraphics[width=0.35\textwidth]{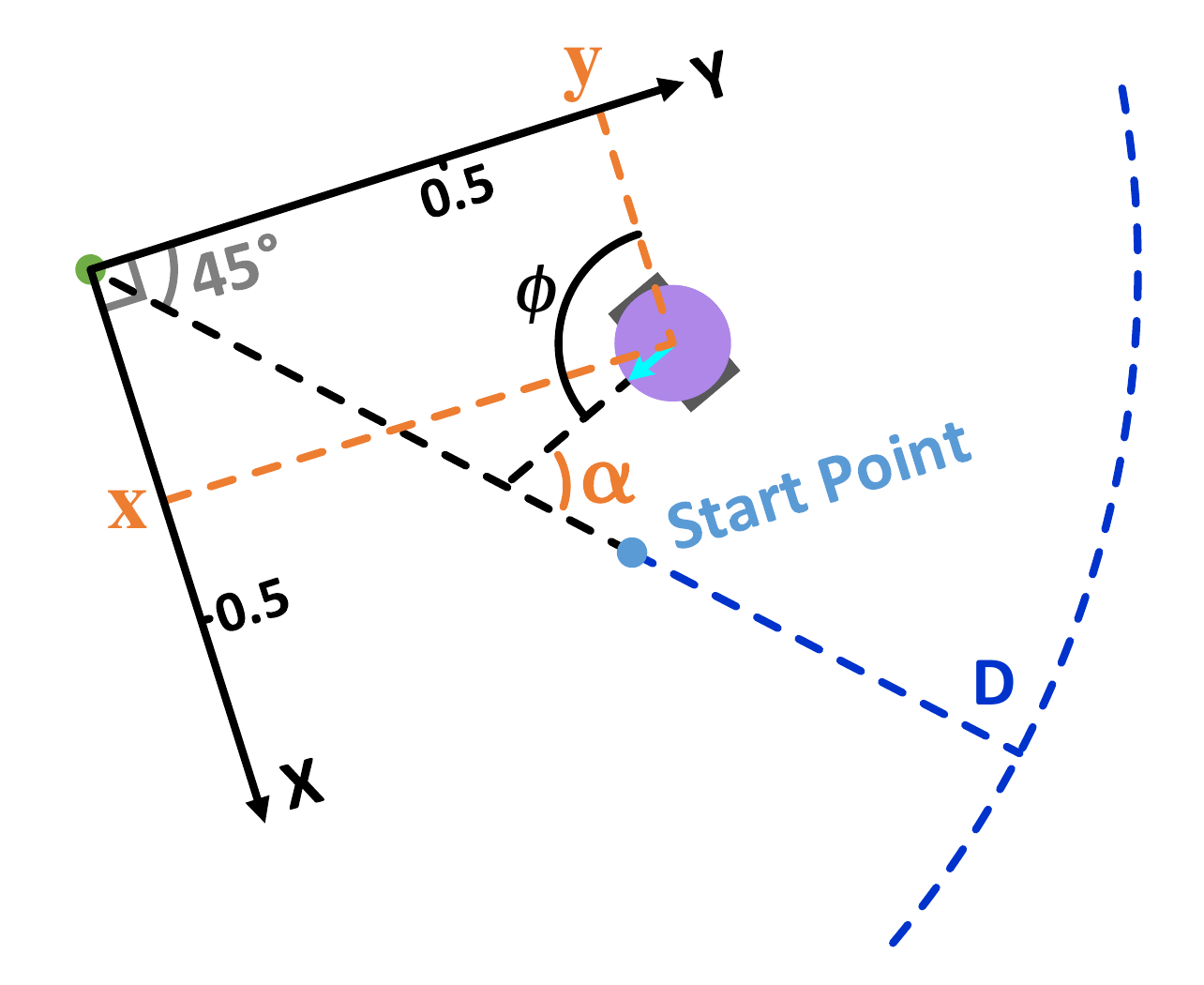}
		\label{s_eval}}
	\caption{Normalized relative state representation of Sparrow. Here, $L$ is the training map size, and $D$ is the maximum local planning distance.}
	\label{s_r}
\end{figure}

\noindent \textbf{State}: The state of the agent is a vector of length 32, containing the pose of the robot ($x$, $y$ and $\alpha$ as shown in Fig. \ref{s_r} (a)), the linear and angular velocity of the robot ($v_l$ and $v_a$), and 27 scanning results of the LiDAR mounted on the robot. The state variables will be normalized and represented in a relative fashion before being fed to the agent. The normalized relative state representation could simplify the training: we could train the agent in a fixed manner (start from the lower left corner, and end at the upper right corner), and the trained agent is capable of handling any start-end scenarios as long as their distance is within $D$, as illustrated in Fig. \ref{s_r} (b).

\noindent \textbf{Action}: We employ 5 discrete actions to control the target velocity $ [\hat{v}^{t}_{l},\hat{v}^{t}_{a}]$ of the robot, which are:

\begin{itemize}
	\item{\textit{Turn left}: [0.36 cm/s, 1 rad/s]}
	\item{\textit{Go straight and turn left}:[18 cm/s, 1 rad/s]}
	\item{\textit{Go straight}: [18 cm/s, 0 rad/s]}
	\item{\textit{Go straight and turn right}: [18 cm/s, -1 rad/s]}
	\item{\textit{Turn right}: [0.36 cm/s, -1 rad/s]}
\end{itemize}

\noindent \textbf{Reward}: If the robot collides with an obstacle or reaches the end point, the episode is terminated, and the agent is rewarded with -10 or 75, respectively. Otherwise, the agent receives a reward according to:
\begin{equation}
	\label{rf}
	r=0.3 r_{d 1}+0.1 r_{d 2}+0.3 r_v+0.3 r_\alpha+0.1 r_d
\end{equation}
\noindent where $r_{d 1}$ and $r_{d 2}$  are negatively correlated to $d1$ and $d2$ (see Fig. \ref{s_r} (a)), with a maximum value of $1$; $r_v=1$ if the linear velocity of the robot exceeds half of its maximum linear velocity, otherwise 0; $r_\alpha$ is negatively correlated to the absolute value of $\alpha$; $r_d=-1$ if the closest distance between the robot and an obstacle is smaller than 30 cm, otherwise 0. The reward function is devised with the intention of driving the robot to the end point expeditiously without colliding with obstacles. For more implementation details, please refer to our website\footnotemark[1].

\subsubsection{Task2Task generalization}
In this section, we examine the Task2Task generalization capability of the agent trained by Color. All experiments were conducted within the Sparrow simulator.  More specifically, we seek to investigate the extent to which the vectorized diversity of Sparrow contributes to the Task2Task generalization. 

\begin{figure*}[t]
	\centering
	\includegraphics[width=0.8\textwidth]{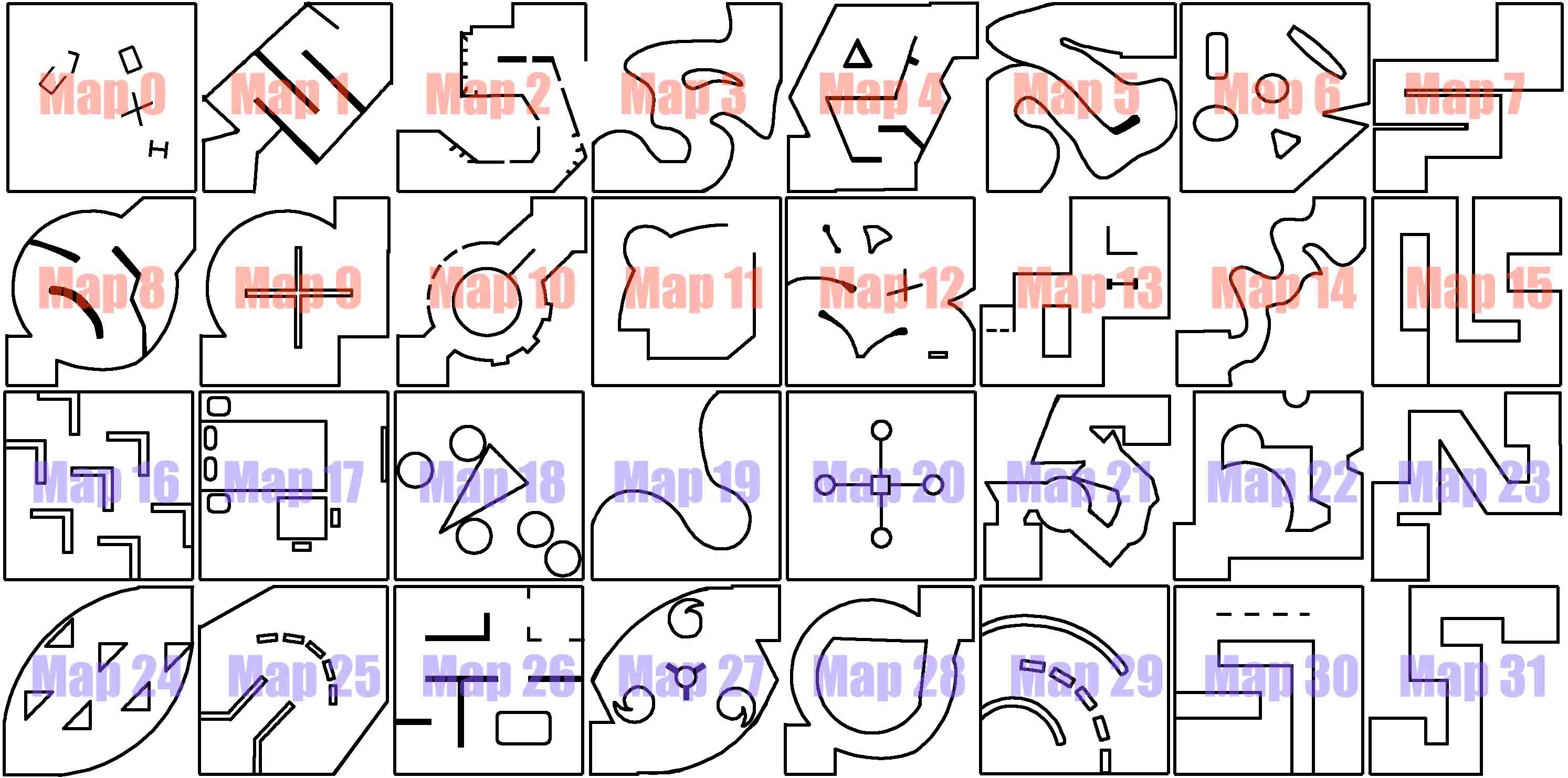}
	\caption{Training (0$\sim$15) and test (16$\sim$31) maps for Task2Task generalization assessment.}
	\label{maps}
\end{figure*}

To this end, two experimental setups have been designed. The first setup, denoted as Grey, employs identical simulation parameters and training maps (Map0$\times$16) across all vectorized environments. In contrast, the other setup, referred to as Color, randomizes the control interval, control delay, velocity range, kinematic parameter $K$, and the magnitude of sensor noise across the vectorized environments.
\color{black}
Note that the initial values of these simulation parameters, as presented in Table \ref{tab:sim_params}, were approximately derived from empirical measurements of our real-world robot, depicted in Fig. \ref{S&R} (b), utilizing a motion capture system\footnote{FZMotion Motion Capture System: https://www.luster3ds.com/}. 
The randomization is realized by uniformly sampling from a moderate interval around the initial values, which is performed at the onset of each episode.
\color{black}
Additionally, a collection of diverse maps is leveraged for training, as the Map0 to Map15 shown in Fig. \ref{maps}. Notably, both Grey and Color employ the ASL training framework.

\begin{figure}[t]
	\centering
	\subfloat{\includegraphics[width=0.45\textwidth]{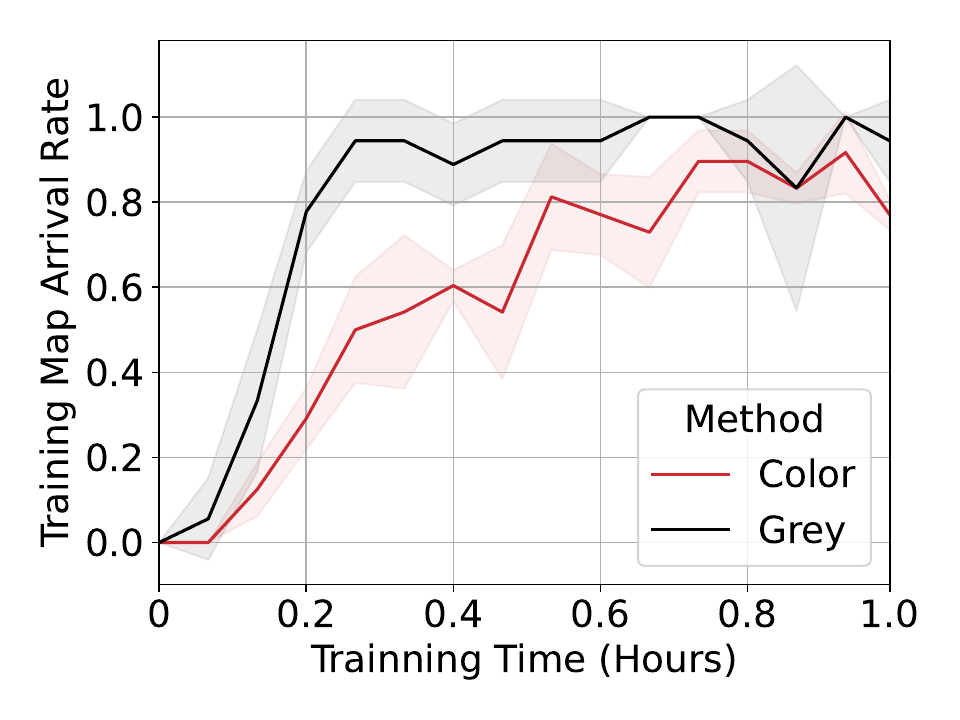}
		\label{train_rate}}
	\hfil
	\subfloat{\includegraphics[width=0.45\textwidth]{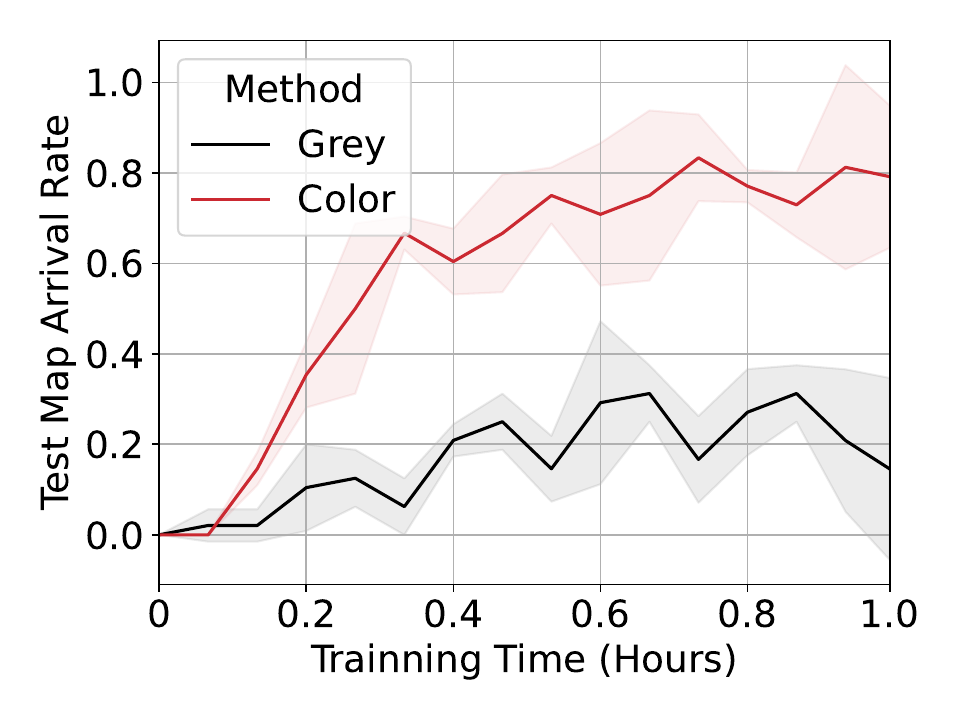}
		\label{test_rate}}
	\caption{\textcolor{black}{Target area arrival rate on training maps (left) and test maps (right). The solid curves represent the mean value across three random seeds, and the translucent areas correspond to the standard derivation. Note that the experiments presented here take only one hour of simulation training.}}
	\label{arrival_rate}
\end{figure}

It is worth noting that, both for Grey and Color, the obstacles in Map0 and the initial pose of the robot are randomly generated to evade overfitting. We harness 4 fully connected layers of shape [32, 256, 128, 5] to map the state variable to the discrete actions. Other hyperparameters are listed in Table \ref{tab:hp}. The Color and Grey were trained until converged under three different random seeds, approximately equating to \textbf{one hour} of wall-clock time per seed. During training, the models of Color and Grey were evaluated on their respective training maps and the common test maps (Map16 to Map31), and the corresponding target area arrival rate curves were recorded in Fig. \ref{arrival_rate}. 

Fig. \ref{arrival_rate} demonstrates that the overfitting, analogous to that in supervised learning, also occurs in DRL. As mirrored by the Grey, the homogeneous environments could result the DRL agent in overfitting and impair its generalization capability. 
\textcolor{black}{This observation is evidenced by: 1) only 30\% arrival rate on test maps 2) the large fluctuation on training maps at approximately 0.9 hours. Specifically, the large fluctuation stems from reduced data diversity following the attainment of 100\% arrival rate on training maps near 0.7 hours. Consequently, Grey overfits ideal navigation trajectories yet struggles to manage atypical situations such as "circling" and "collision" events, as commonly referred to as the out-of-trajectory \citep{OOT} generalization challenge in DRL.}
Impressively, thanks to the diversified environment setups, the Color is able to achieve an arrival rate over 80\% on the zero-shot generalization tests, with a tolerable loss of converge speed and final performance on training maps. These results validly corroborate the beneficial impact of Sparrow's vectorized diversity on Task2Task generalization.

\subsubsection{Sim2Real generalization}

\begin{figure}[H]
	\centering
	 \adjustbox{width=1\textwidth, center}{\includegraphics[width=\textwidth]{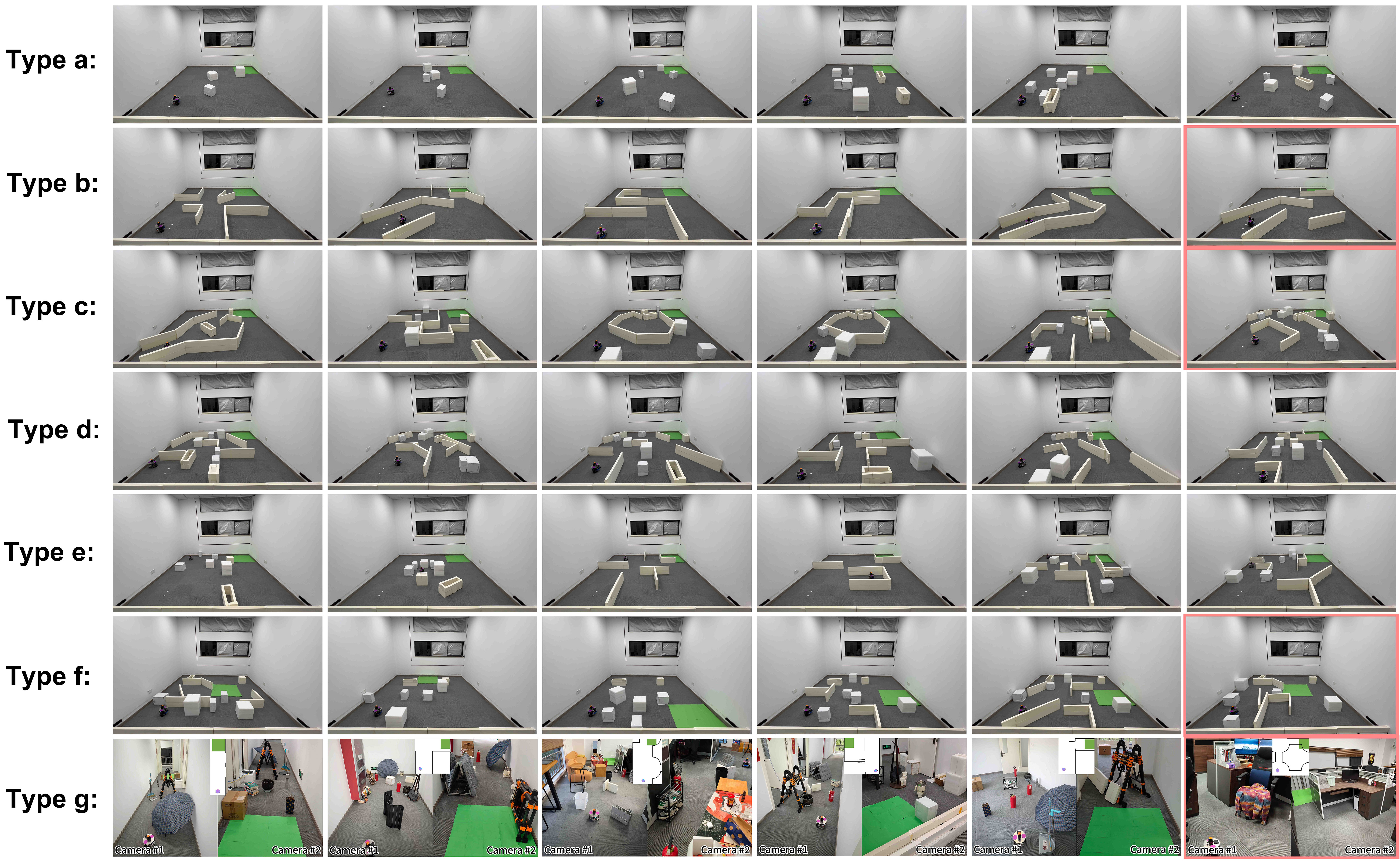}}
	\caption{\textcolor{black}{42 real-world scenarios for Sim2Real assessment: a) Blocky; b) Fence-shaped; c) Simply mixed; d) Intricately mixed; e) Random start point; f) Random end point; g) Complicated Scenarios. The task is to navigate the robot to the green area without collision. Failure cases are marked with red boxes. The video is available on the website}\protect\footnotemark[1].}
	\label{real_all}
\end{figure}

\begin{figure}[H]
	\centering
	\subfloat[]{\includegraphics[width=0.23\textwidth]{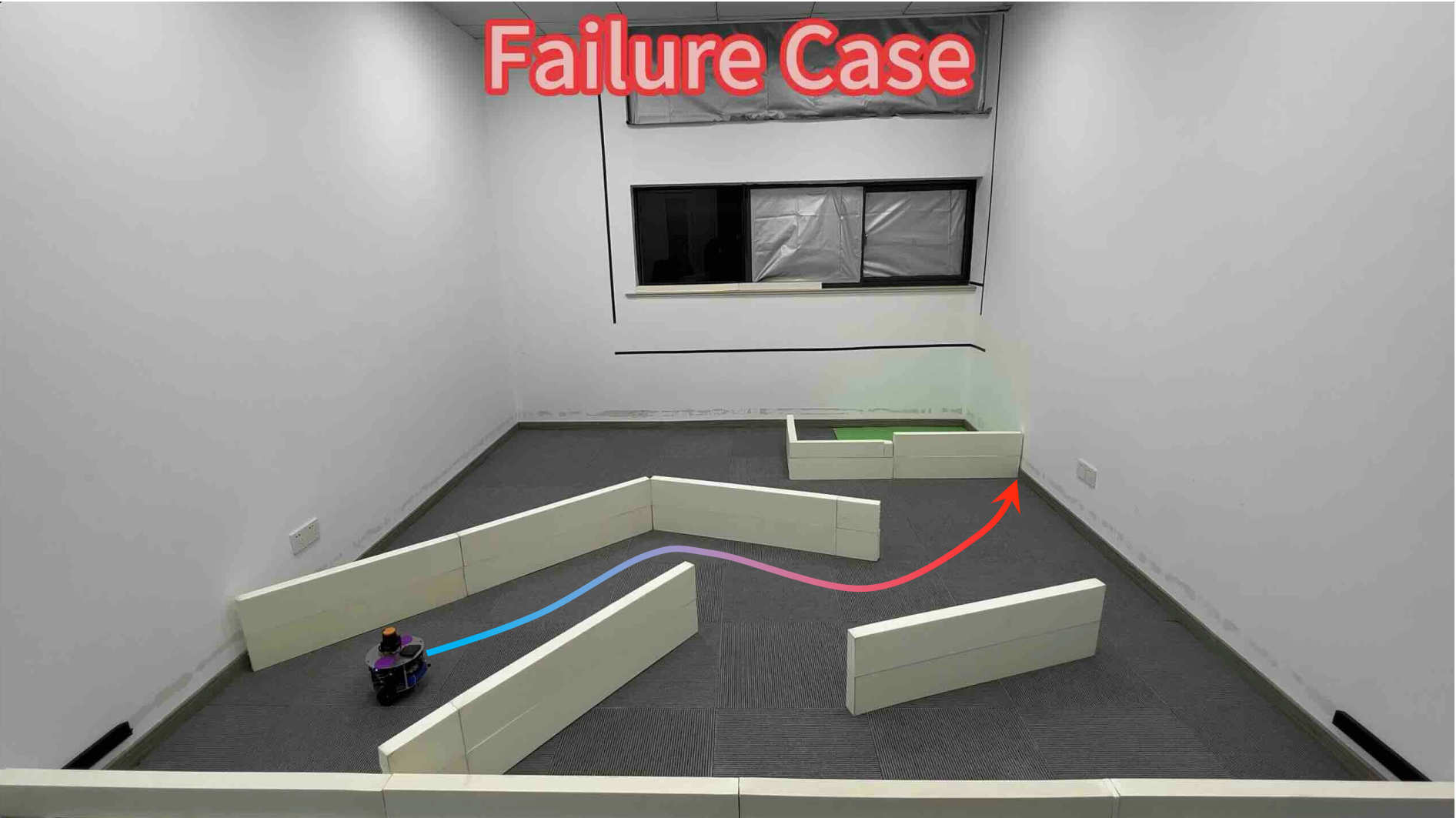}}
	\hfil
	\subfloat[]{\includegraphics[width=0.23\textwidth]{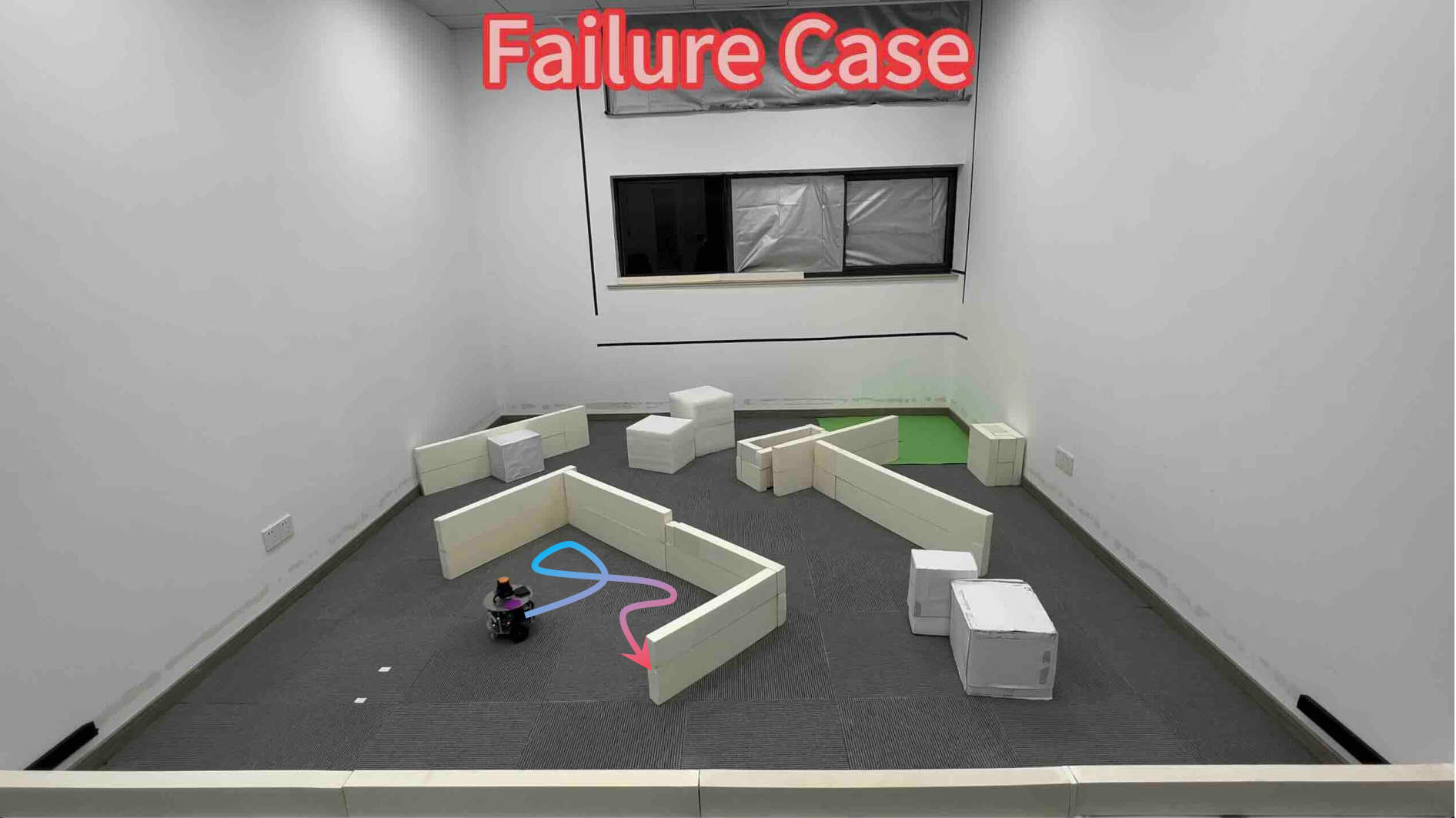}}
	\hfil
	\subfloat[]{\includegraphics[width=0.23\textwidth]{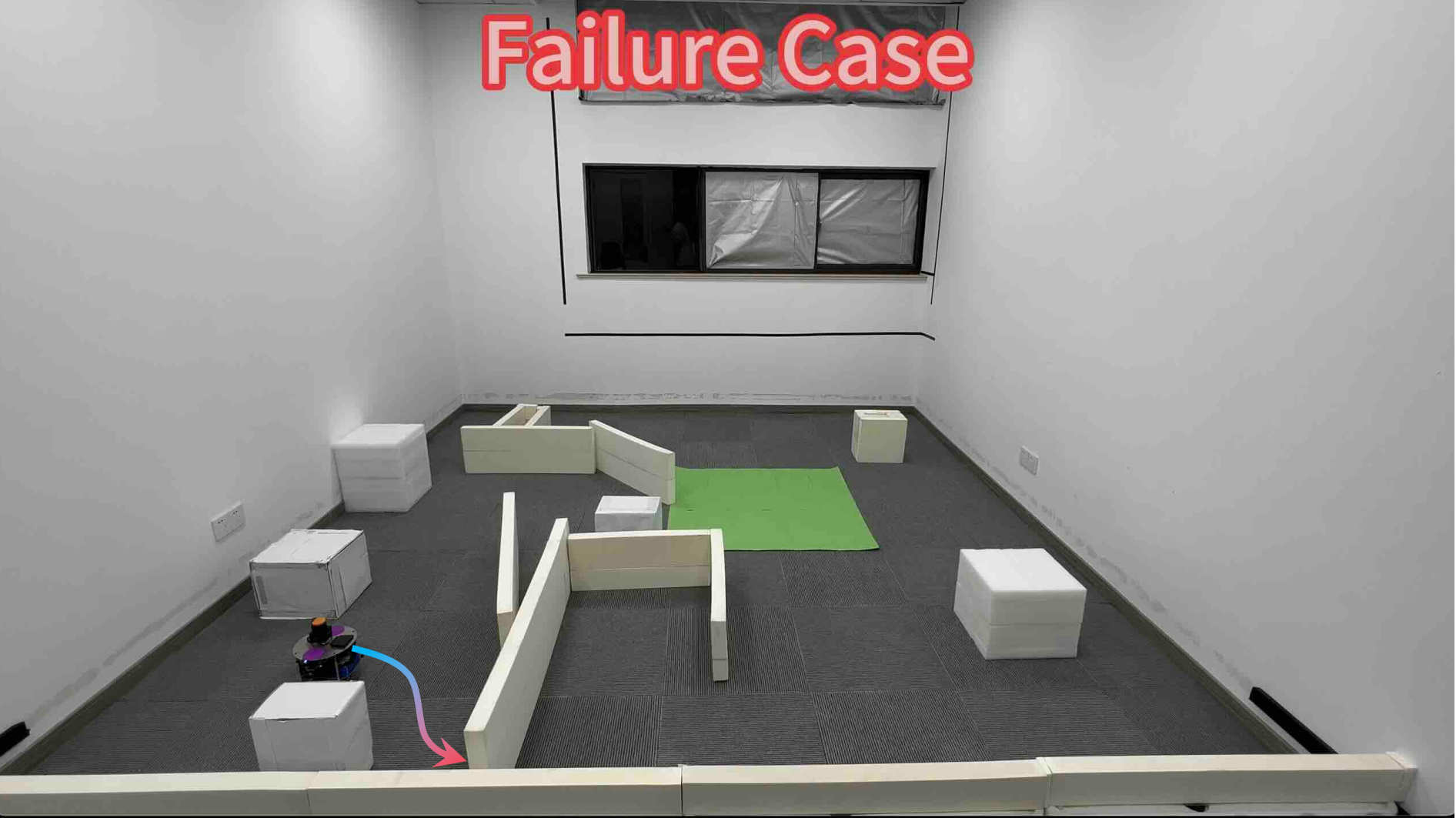}}
 	\hfil
        \subfloat[]{\includegraphics[width=0.23\textwidth]{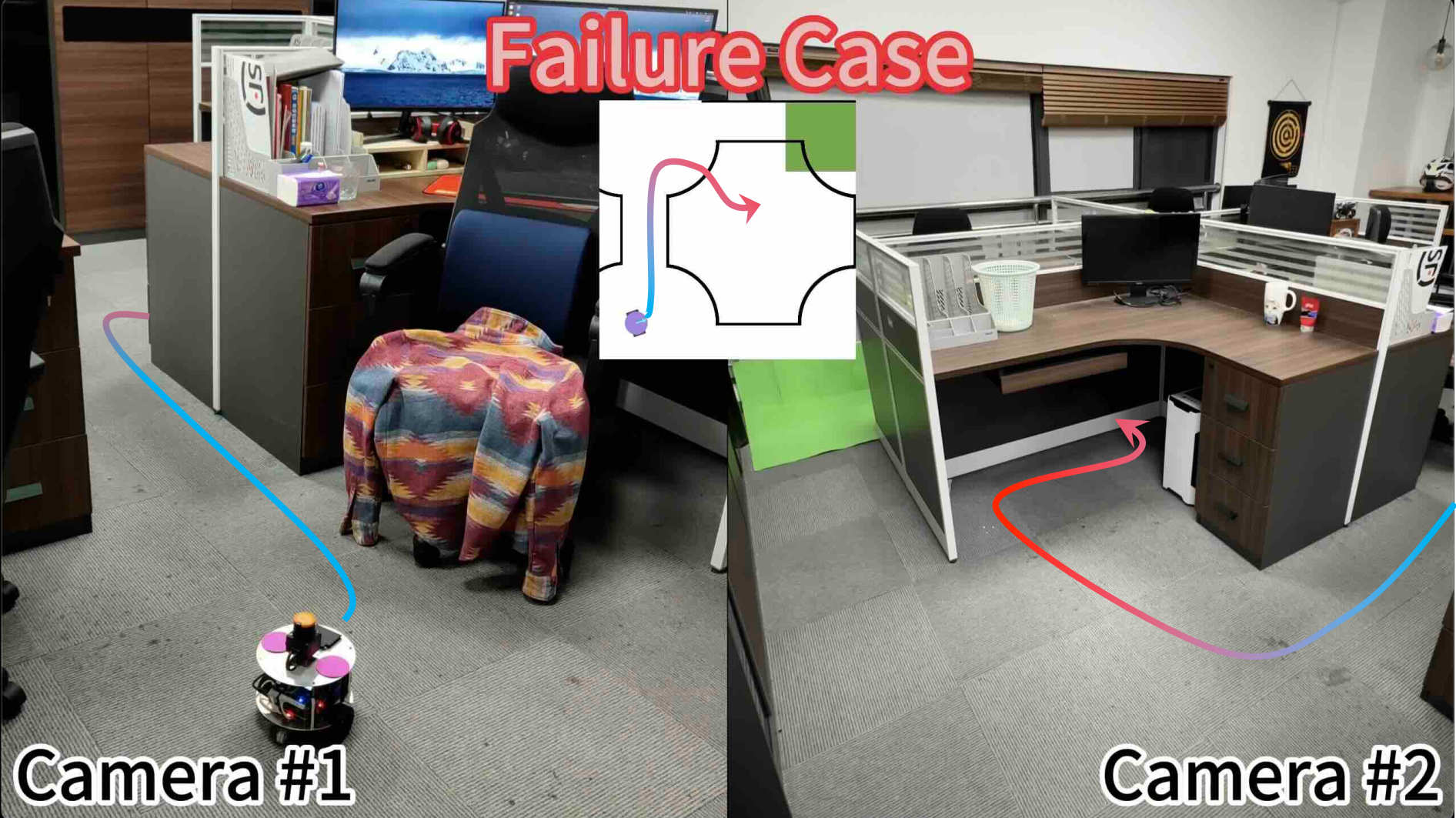}}
	\caption{\textcolor{black}{Failure cases of real-world experiments.}}
	\label{failedcases}
\end{figure}

In this section, 42 real-world scenarios of 7 types have been built to examine the Sim2Real generalization capability of the agent trained by Color, as shown in Fig. \ref{real_all}. We select the best-performing planning model (the trained neural networks) derived from the previous section as the real-world planning model. To demonstrate the superiority of Color, the planning model is directly implanted into the robot (see Fig. \ref{S&R} (b)) to undertake the LPP tasks in the real world without any additional tuning or adaption. Impressively, the agent demonstrates superior proficiency in real-world local path planning, succeeding in navigating to the end point in 38 out of 42 total scenarios. The four failure cases have been designated in red within Fig. \ref{real_all}. 
\color{black}
Fig. \ref{failedcases} and the experiment video\footnotemark[1] reveal that these failures predominantly result from the robot's attempts to forcefully navigate through obstacles. This phenomenon is primarily attributed to the partially observable nature of the LPP problem, as the Color agent lacks ground truth knowledge of the maps and is thus susceptible to entrapment in snares. Three potential approaches to address this issue are: 
(i) Equipping Color with map information via integration with GPP algorithms; 
(ii) Endowing Color with memory capabilities by employing temporal awareness networks; 
(iii) Improving Color's generalization capacity through the design of more appropriate reward functions and training maps.
Due to the limited scope of this paper, further investigations into these approaches cannot be conducted at present but will be the focus of future research endeavors.

\color{black}
Compared with other solutions concerning the real-world application of DRL in the context of local planning or navigation, the advantages of Color can be more intuitively sensed. 
Unlike \citet{realworld_human}, which necessitates human guidance for fine-tuning the agent in real-world scenarios, our method operates autonomously without human intervention or fine-tuning, enhancing both efficiency and simplicity.
In \citet{realworld_30hour}, due to the scarcity of parallel simulation, approximately 30 hours of simulation training are required to produce a planner compatible with real-world applications. Furthermore, its Sim2Real transfer does not account for the robot's dynamics, constraining its movement to a discretized grid world. In contrast, Sparrow's parallel simulation capability, combined with the high efficiency of the ASL framework, enables the training of a real-world planner in merely one hour of simulation. Additionally, the vectorized diversity of Sparrow addresses the dynamics issue, ensuring continuous robot motion in real-world tasks.
In \citet{realworld_nature}, a mathematical model of the robot and a residual model of the Sim2Real discrepancy are required to ensure the Sim2Real transfer. In contrast, our approach is environmentally model-free. Devoid of any precise information about the robot and environments, Color is capable of managing a variety of local planning tasks, both in simulation and the real world.
\color{black}

\color{black}
\subsection{Ablation studies}
We conducted ablation studies to determine the respective contributions of ASL and Sparrow, as illustrated in Table \ref{ablation_result}.

\begin{table}[H]
    \begin{adjustbox}{center}
    \resizebox{\textwidth}{!}{
    \begin{tabular}{cccccccc}
        \hline
        \multirow{2}[0]{*}{\textbf{Case}}   & \multirow{2}[0]{*}{\textbf{Ablated Part}} & \multicolumn{2}{c}{\textbf{Experiment Setups}} & \multirow{2}[0]{*}{\textbf{0.2TAR-Samples}} & \multirow{2}[0]{*}{\textbf{0.2TAR-Time}} & \multirow{2}[0]{*}{\textbf{TAR}} & \multirow{2}[0]{*}{\textbf{RAR}} \\
        \cline{3-4}
        & & \textbf{DRL Algorithm} & \textbf{Simulator} & & & & \\
        \hline
        1 & - & ASL DDQN & V-Sparrow with VD & 0.22 M & 0.14 H & 84\% & 90\% \\

        2 & ASL & Vectorized DDQN & V-Sparrow with VD & 0.30 M & 0.26 H & 79\% & 83\% \\

        3 & Sparrow & ASL DDQN & V-Sparrow without VD  & 0.62 M & 0.40 H & 25\% & 19\% \\

        4 & ASL+Sparrow & DDQN & NV-Sparrow without VD  & 0.71 M & 1.16 H & 22\% & 19\% \\
        \hline
        \multicolumn{8}{l}{V: Vectorized; NV: Non-Vectorized; VD: Vectorized Diversity; M: Million samples; H: Hours}\\
    \end{tabular}}
    \end{adjustbox}
    \caption{Results of ablation studies}
    \label{ablation_result}
\end{table}

\subsubsection{Ablation setups}

Here, case 1 represents Color.
Case 2 is designed to ablate the ASL framework. Here, as the Sparrow simulator is characterized by vectorized diversity, which relies on vectorization, the vectorized interaction mode is maintained to ensure compatibility between the ablated DRL algorithm and Sparrow. Other components of ASL, such as VEM and the partially decoupled framework, are removed, resulting in the vectorized DDQN algorithm. 
Case 3 is designed to ablate Sparrow. Similar to Case 2, the ASL framework also relies on vectorization. Therefore, we ablate the main feature of Sparrow, namely vectorized diversity, while retaining its vectorized simulation with identical environmental copies.
Case 4 serves as a baseline, with both ASL and Sparrow ablated.
It is important to note that for cases without vectorized diversity, Map0 in Fig. \ref{maps} is employed as the training map, with obstacles randomly generated for each interaction episode to mitigate overfitting.

\subsubsection{Evaluation metrics}
We established four metrics to evaluate the generalization capacity and efficiency of the aforementioned ablation cases:

\noindent\textbf{RAR}: \textbf{R}eal maps (Fig. \ref{real_all}) \textbf{a}rrival \textbf{r}ate, designed to assess Sim2Real generalization.

\noindent\textbf{TAR}: \textbf{T}est maps (Fig. \ref{maps}) \textbf{a}rrival \textbf{r}ate, designed to evaluate Task2Task generalization.

\noindent\textbf{0.2TAR-Sample}: The number of training samples required to achieve a 20\% arrival rate on test maps, designed to measure sample efficiency.

\noindent\textbf{0.2TAR-Time}: The wall-clock training time required to achieve a 20\% arrival rate on test maps, designed to determine training time efficiency.

The 0.2 threshold was selected to ensure accessibility, as it represents the lower bound achievable by all cases.

\subsubsection{Analysis of the ablation results}
The ablation results are presented in Table \ref{ablation_result}. Note that these results are averaged over three independent trials. 
The ASL training framework predominantly contributes to sample and training time efficiency, as evidenced by the comparison between cases 1 and 2, as well as the comparison between cases 3 and 4. The enhanced sample efficiency can be attributed to ASL's VEM, which achieves a superior balance between exploration and exploitation, thereby improving training sample quality and elevating efficiency. The improvement in training time efficiency is attributable to ASL's VDC and partially decoupled training mode. VDC operates in a batched fashion to accelerate data collection, while the partially decoupled training mode reduces training time by executing data collection and model training concurrently.

The Sparrow simulator demonstrates notable effectiveness in enhancing generalization capacity, as indicated by the comparison between cases 1 and 3. This enhancement stems from the data diversity provided by Sparrow, which prevents the agent from overfitting to a fixed map layout (Task2Task) and robot characteristics (Sim2Real). Notably, Sparrow also significantly improves sample and training time efficiency. We believe such improvement is derived from Sparrow’s vectorized diversity as well, where the vectorization promotes data throughput, and the diversity enhances data quality.
\color{black}

\section{Conclusion}
\label{section5}

\textcolor{black}{This paper addresses the deficient training efficiency and generalization capability of DRL, proposing a DRL-based solution to the LPP problem named Color.} Specifically, to improve the efficiency of the DRL algorithms, a partially decoupled training framework ASL is proposed. It is noteworthy that the ASL is specifically tailored to accommodate off-policy DRL algorithms that utilize experience replay. DRL algorithms that satisfy this requirement are well-suited for integration with ASL to enjoy a further promotion in sample and training time efficiency. Furthermore, a Sparrow simulator supporting vectorized diversity is developed to enhance both the training efficiency and generalization capability. Sparrow and ASL are then integrated via their interdependent vectorized environments, culminating in the Color system. Impressively, with one hour of simulation training, the Color successfully completed 38 out of 42 real-world LPP tasks, which we believe have reached a new milestone in the real-world application of DRL. \textcolor{black}{Despite promising results, this study has limitations, including the extension of Color to dynamic environments. This extension necessitates equipping Sparrow with dynamic obstacle simulation capabilities and integrating temporal perception learning into ASL, which constitute the main focus of our future research. Another prospective research direction involves Color’s integration with global path planning algorithms, potentially expanding Color's applicability in scenarios such as mall service robots and warehouse transportation systems.}

\section*{Acknowledgement} 
We acknowledge the support from the National Natural Science Foundation of China under Grant No. 62273230 and 62203302. 

We also acknowledge the support from the Ministry of Education, Singapore, under its Academic Research Fund Tier 1 (RG136/22). Any opinions, findings and conclusions or recommendations expressed in this material are those of the authors and do not reflect the views of the Ministry of Education, Singapore.


\section*{Appendix}

\begin{table}[htbp]
\small
\center
\caption{\textcolor{black}{Hyperparameters}}
\label{tab:hp}%
\begin{minipage}{0.45\textwidth}
\begin{tabular}{lll}
		\hline
		\textbf{Framework-associated} & \textbf{Atari} & \textbf{Sparrow} \\
		\hline
		$N$ & 128 & 16 \\
		$\mbox{\textit{TPS}}$ & 8 & 256 \\
		Linear Decay Steps of $OR$ & 500K $T_{step}$ & 500K $T_{step}$ \\
		$OR$ & 128 $\rightarrow$ 4 & 16 $\rightarrow$ 3 \\
		$E_{min}$ & 0.01 & 0.01 \\
		$E_{max}$ & 0.8 & 0.8 \\
		$U$ & 50 $B_{step}$ & 50 $B_{step}$ \\
		\hline
	\end{tabular}%
\end{minipage}
\hfill
\begin{minipage}{0.45\textwidth}
\begin{tabular}{lll}
		\hline
		\textbf{Algorithm-associated} & \textbf{Atari} & \textbf{Sparrow} \\
		\hline
		Learning Start Steps $C$ & 150K transitions & 30K transitions\\
		Replay Buffer Size & 1M transitions & 1M transitions \\
		$\gamma$ & 0.99 & 0.98 \\
		Learning rate & $6.25 \times 10^{-5}$ & $1.0 \times 10^{-4}$ \\
		Target Net Update Frequency & 2k $B_{step}$ & 200 $B_{step}$ \\
		Mini-batch size & 32 & 256 \\
		Optimizer & Adam  & Adam \\
		\hline
	\end{tabular}%
\end{minipage}
\end{table}

\begin{table}[htbp]
\small
\center
\begin{minipage}{0.45\textwidth}
\caption{\textcolor{black}{Values of simulation parameters}}
\label{tab:sim_params}%
\begin{tabular}{p{0.6\textwidth}p{0.4\textwidth}}
\hline
\textbf{Parameter} & \textbf{Value}  \\
\hline
$K$&0.6\\
Control interval&0.1 s\\
Control delay&0.1 s \\
Maximal linear velocity&18 cm/s\\
Maximal angular velocity&1 rad/s\\
Magnitude of sensor noise&1 cm\\
\hline
\end{tabular}%
\end{minipage}
\hfill
\begin{minipage}{0.45\textwidth}
\caption{\textcolor{black}{Hardware Platform}}
\label{tab:hardware}%
\begin{tabular}{p{0.4\textwidth}p{0.6\textwidth}}
\hline
\textbf{Component} & \textbf{Description}  \\
\hline
CPU&AMD 3990X\\
GPU&Nvidia RTX 3090\\
SSD&Samsumg 990 Pro 2TB \\
RAM&Kingston 32GB 5600MHz × 2\\
Mobile Robot&WHEELTEC-Differential\\
LiDAR&HOKUYO-UST-10LX\\
\hline
\end{tabular}%
\end{minipage}
\end{table}

\begin{table}[H]
\small
\center
\caption{\textcolor{black}{Software Platform}}
\label{tab:software}%
\begin{tabular}{p{0.25\textwidth}p{0.25\textwidth}}
\hline
\textbf{Component} & \textbf{Description}  \\
\hline
Computer System&Ubuntu 20.04.1\\
Robot System&ROS noetic\\
Programming Language&Python 3.8.0\\
\hline
\end{tabular}%
\end{table}%

\begin{table}[H]
\small
\center
\caption{\textcolor{black}{Python Dependencies}}
\label{tab:pythonlib}%
\begin{tabular}{p{0.25\textwidth}p{0.25\textwidth}}
\hline
\textbf{Libraries} & \textbf{Version}  \\
\hline
torch&2.0.1\\
numpy&1.24.3\\
pygame&2.4.0\\
envpool&0.6.6\\
\hline
\end{tabular}%
\end{table}%

\newpage

\newpage

\bibliographystyle{elsarticle-harv} 
\bibliography{color}

\end{document}